\documentclass{amsart}

\makeatletter

\usepackage{amsmath,amssymb,amsthm,mathtools}
\usepackage{booktabs,array,multirow}
\usepackage{graphicx}
\usepackage{xcolor}
\usepackage{enumitem}
\usepackage{microtype}
\usepackage{url}

\@ifundefined{algorithm}{\usepackage{algorithm}}{}
\@ifundefined{State}{\usepackage{algpseudocode}}{}

\@ifpackageloaded{hyperref}{}{%
  \usepackage[colorlinks=true,
              linkcolor=blue!45!black,
              citecolor=blue!45!black,
              urlcolor=blue!45!black]{hyperref}}

\usepackage[initials,backrefs]{amsrefs}

\@ifpackageloaded{cleveref}{}{%
  \usepackage[capitalise,nameinlink]{cleveref}}

\newcommand\project@newtheorem[3]{%
  \@ifundefined{#1}{\newtheorem{#1}[#2]{#3}}{}}

\@ifundefined{theorem}{\newtheorem{theorem}{Theorem}[section]}{}
\project@newtheorem{lemma}{theorem}{Lemma}
\project@newtheorem{proposition}{theorem}{Proposition}
\project@newtheorem{corollary}{theorem}{Corollary}
\project@newtheorem{conjecture}{theorem}{Conjecture}
\theoremstyle{definition}
\project@newtheorem{definition}{theorem}{Definition}
\project@newtheorem{example}{theorem}{Example}
\project@newtheorem{assumption}{theorem}{Assumption}
\project@newtheorem{hypothesis}{theorem}{Hypothesis}
\project@newtheorem{problem}{theorem}{Problem}
\project@newtheorem{question}{theorem}{Question}
\project@newtheorem{notation}{theorem}{Notation}
\theoremstyle{remark}
\project@newtheorem{remark}{theorem}{Remark}
\project@newtheorem{observation}{theorem}{Observation}
\theoremstyle{plain}

\project@newtheorem{thm}{theorem}{Theorem}
\project@newtheorem{lem}{theorem}{Lemma}
\project@newtheorem{prop}{theorem}{Proposition}
\project@newtheorem{cor}{theorem}{Corollary}
\project@newtheorem{conj}{theorem}{Conjecture}
\theoremstyle{definition}
\project@newtheorem{defn}{theorem}{Definition}
\project@newtheorem{exa}{theorem}{Example}
\project@newtheorem{assum}{theorem}{Assumption}
\project@newtheorem{hyp}{theorem}{Hypothesis}
\theoremstyle{remark}
\project@newtheorem{rem}{theorem}{Remark}
\theoremstyle{plain}

\@ifundefined{mainthm}{%
  }{}

\crefname{theorem}{Thm.}{Thms.}                \Crefname{theorem}{Thm.}{Thms.}
\crefname{thm}{Thm.}{Thms.}                    \Crefname{thm}{Thm.}{Thms.}
\crefname{lemma}{Lem.}{Lems.}                  \Crefname{lemma}{Lem.}{Lems.}
\crefname{lem}{Lem.}{Lems.}                    \Crefname{lem}{Lem.}{Lems.}
\crefname{proposition}{Prop.}{Props.}          \Crefname{proposition}{Prop.}{Props.}
\crefname{prop}{Prop.}{Props.}                 \Crefname{prop}{Prop.}{Props.}
\crefname{corollary}{Cor.}{Cors.}              \Crefname{corollary}{Cor.}{Cors.}
\crefname{cor}{Cor.}{Cors.}                    \Crefname{cor}{Cor.}{Cors.}
\crefname{conjecture}{Conj.}{Conjs.}           \Crefname{conjecture}{Conj.}{Conjs.}
\crefname{conj}{Conj.}{Conjs.}                 \Crefname{conj}{Conj.}{Conjs.}
\crefname{definition}{Def.}{Defs.}             \Crefname{definition}{Def.}{Defs.}
\crefname{defn}{Def.}{Defs.}                   \Crefname{defn}{Def.}{Defs.}
\crefname{example}{Ex.}{Exs.}                  \Crefname{example}{Ex.}{Exs.}
\crefname{exa}{Ex.}{Exs.}                      \Crefname{exa}{Ex.}{Exs.}
\crefname{assumption}{Assum.}{Assums.}         \Crefname{assumption}{Assum.}{Assums.}
\crefname{assum}{Assum.}{Assums.}              \Crefname{assum}{Assum.}{Assums.}
\crefname{hypothesis}{Hyp.}{Hyps.}             \Crefname{hypothesis}{Hyp.}{Hyps.}
\crefname{hyp}{Hyp.}{Hyps.}                    \Crefname{hyp}{Hyp.}{Hyps.}
\crefname{problem}{Prob.}{Probs.}              \Crefname{problem}{Prob.}{Probs.}
\crefname{question}{Ques.}{Ques.}              \Crefname{question}{Ques.}{Ques.}
\crefname{notation}{Not.}{Nots.}               \Crefname{notation}{Not.}{Nots.}
\crefname{remark}{Rem.}{Rems.}                 \Crefname{remark}{Rem.}{Rems.}
\crefname{rem}{Rem.}{Rems.}                    \Crefname{rem}{Rem.}{Rems.}
\crefname{mainthm}{Theorem}{Theorems}          \Crefname{mainthm}{Theorem}{Theorems}
\crefname{observation}{Obs.}{Obs.}             \Crefname{observation}{Obs.}{Obs.}
\crefname{algorithm}{Alg.}{Algs.}              \Crefname{algorithm}{Alg.}{Algs.}
\crefname{equation}{Eq.}{Eqs.}                 \Crefname{equation}{Eq.}{Eqs.}
\crefname{section}{Sec.}{Secs.}                \Crefname{section}{Sec.}{Secs.}
\crefname{subsection}{Sec.}{Secs.}             \Crefname{subsection}{Sec.}{Secs.}
\crefname{figure}{Fig.}{Figs.}                 \Crefname{figure}{Fig.}{Figs.}
\crefname{table}{Tab.}{Tabs.}                  \Crefname{table}{Tab.}{Tabs.}
\crefname{appendix}{App.}{Apps.}               \Crefname{appendix}{App.}{Apps.}

\creflabelformat{equation}{#2\textup{(#1)}#3}

\newcommand\ProvideMathOperator[2]{%
  \@ifundefined{\expandafter\@gobble\string#1}{\DeclareMathOperator{#1}{#2}}{}}

\providecommand{\R}{\mathbb{R}}

\providecommand{\SO}{\mathrm{SO}}

\providecommand{\SL}{\mathrm{SL}}

\providecommand{\GL}{\mathrm{GL}}

\providecommand{\abs}[1]{\lvert #1\rvert}
\providecommand{\norm}[1]{\lVert #1\rVert}
\providecommand{\ip}[2]{\langle #1,#2\rangle}

\ProvideMathOperator{\tr}{tr}
\ProvideMathOperator{\rank}{rank}
\ProvideMathOperator{\diag}{diag}
\ProvideMathOperator{\dist}{dist}
\ProvideMathOperator{\disc}{disc}
\ProvideMathOperator{\Disc}{Disc}
\ProvideMathOperator{\Res}{Res}
\ProvideMathOperator{\lead}{lead}
\ProvideMathOperator{\Fix}{Fix}
\ProvideMathOperator{\Stab}{Stab}
\ProvideMathOperator{\Aut}{Aut}
\ProvideMathOperator{\End}{End}
\ProvideMathOperator{\Hom}{Hom}
\ProvideMathOperator{\Sym}{Sym}
\ProvideMathOperator{\Spec}{Spec}
\ProvideMathOperator{\Jac}{Jac}
\ProvideMathOperator{\Kum}{Kum}
\ProvideMathOperator{\Hess}{Hess}
\ProvideMathOperator{\Gr}{Gr}
\ProvideMathOperator{\Sing}{Sing}
\ProvideMathOperator{\supp}{supp}
\ProvideMathOperator{\sgn}{sgn}
\ProvideMathOperator{\codim}{codim}
\ProvideMathOperator{\Span}{Span}
\ProvideMathOperator{\Id}{Id}
\ProvideMathOperator{\AUC}{AUC}
\ProvideMathOperator{\Emb}{Emb}
\ProvideMathOperator{\Diff}{Diff}
\ProvideMathOperator{\Gram}{Gram}

\AtBeginDocument{%
  \providecommand{\Gm}{\mathbb{G}_m}
  \providecommand{\bq}{\mathbf{q}}
  \providecommand{\btheta}{\boldsymbol{\theta}}
  \providecommand{\Dil}{D}
  \providecommand{\Rot}{R}
  \providecommand{\rhoq}{\rho_{\bq}}
  \providecommand{\rhot}{\rho_{\btheta}}%
  \providecommand{\Hfreq}{\mathcal{H}}
  \providecommand{\GCN}{\textup{GCN}}
  \providecommand{\UCN}{\textup{UCN}}
  \providecommand{\SNT}{\textup{SNT}}
  
  \providecommand{\logsumexp}{\operatorname{logsumexp}}%
  \providecommand{\ReLU}{\operatorname{ReLU}}%
  \providecommand{\Var}{\operatorname{Var}}%
  \providecommand{\Risk}{\operatorname{R}}%
  \providecommand{\Lap}{\operatorname{\Delta}}%
}

\makeatother

\usepackage[margin=1.2in]{geometry}

\begin{document}

\title[Graded convolutional networks]{Graded convolutional networks on Gaussian image jets}

\author[T. Shaska]{T. Shaska}
\address{Department of Computer Science \& Engineering, Oakland University, Rochester, MI 48309}
\email{shaska@oakland.edu}



\date{\today}

\begin{abstract}
We construct graded convolutional networks on Gaussian image jets, with derivative order as the grade. Retraction onto the weighted sphere gives a complete dilation invariant of the nonzero jet. Gaussian blur shifts the smoothing scale for every feature computed from the raw jet. Normalization at the nominal scale introduces an additional factor depending on derivative order. We construct two smooth image families that an ideal graded classifier separates with a uniform positive margin. The score gap tends to zero for radial probe classifiers and for classifiers continuous at the common limit. Fixed positive regularization destroys this margin. Pointwise graded networks approximate continuous dilation- and rotation-invariant functions uniformly on compact sets of nonzero jets. In a reimplementation of Lindeberg's Gaussian derivative network, retracting the first-layer jet raises MNIST accuracy from $91.45\%$ to $95.10\%$ under Gaussian blur of standard deviation two pixels.
\end{abstract}

\maketitle

\section{Introduction}\label{sec-intro}

When an image is enlarged by a factor of two, its gradient at the corresponding Gaussian scale is halved and its Hessian is divided by four. A network that combines first and second derivatives must account for these different responses to the same enlargement. Lindeberg's scale-normalized derivatives do this using the smoothing scale \cite{Lindeberg1998, Lindeberg2022}. We construct a graded convolutional network whose local normalization is computed from the derivatives themselves. The difference matters when the image has already been blurred by an unknown amount.

Let $I$ be an image on $\R^2$ and $L(x;s)=(g_s*I)(x)$ its Gaussian scale space. The derivatives of order $k$ of $L$ at a position $x$ and a scale $s$ are the coefficients of a binary form $f_k$ of degree $k$, and the jet $f=(f_1,\dots,f_N)$ lies in the graded vector space $V^+=\bigoplus_{k=1}^N\Sym^k(\R^2)^*$. A magnification of the image by $\lambda>0$ carries the jet at $(x;s)$ to the jet $\lambda^{-1}\star f$ at $(\lambda x;\lambda^2s)$, where $\mu\star f=(\mu^kf_k)_k$ is the grading action of the multiplicative group \cite{ShaskaGNN2025}: the derivative order is the grade. Derivatives are the natural inputs of a network that respects this action, since the local linear operators on graded fields that commute with translations and dilations are exactly the homogeneous differential operators with constant coefficients, the order being the shift of grade (\cref{prop-gradedconv}). Every network that reads an image through its Gaussian jets carries the action, from the Gaussian derivative networks of Lindeberg \cite{Lindeberg2022, JanssonLindeberg2022, PerzanowskiLindeberg2025, PerzanowskiLindeberg2026} to the networks on harmonic coordinates \cite{Yarotsky2022, JennerWeiler2022}.

There are two ways to remove the action. The first uses the smoothing scale: the \emph{scale-normalized jet} $\nu(f,s)=s^{1/2}\star f$, with the blocks $s^{k/2}f_k$, is the $\gamma$-normalization of \cite{Lindeberg1998} with $\gamma=1$. The second uses the jet alone: the \emph{graded retraction}
\[
\pi_{\bq}(f)=\Bigl(\frac{f_k}{\rhoq(f)^k}\Bigr)_{k=1}^N,\qquad \rhoq(f)=\Bigl(\sum_{k=1}^N\norm{f_k}^{2/k}\Bigr)^{1/2},
\]
divides the block of order $k$ by the $k$th power of a homogeneous norm of the jet, and maps $V^+\smallsetminus0$ onto the weighted sphere $S_{\bq}=\{\rhoq=1\}$ along the orbits of the grading action. Our first result is that these two maps account for all invariant local features (\cref{thm-forced}). A feature of the jet and the scale is invariant under magnification if and only if it is a function of the scale-normalized jet, and a feature of the nonzero jet alone is invariant if and only if it is a function of its retraction. The grades are forced: normalizing the orders with weights other than $1,\dots,N$, or retracting onto the weighted sphere of weights not proportional to them, is not invariant. The second construction discards one number that the first keeps, the scale-normalized energy $s^{1/2}\rhoq(f)$ of the jet. Discarding it has a price, since this number says how strong the local structure is at its scale, and a use, since it is the only place where the smoothing scale enters: a feature built on the retraction does not depend on a scale parameter that may be wrong.

The scale parameter is wrong when the image arrives already smoothed, by defocus or by a reduction of resolution. A Gaussian blur of variance $b$ before the network makes the jet observed at the scale $s$ equal to the jet of the original image at the scale $s+b$. Every function of the raw jet inherits this identity, the ordinary radial projection included. The scale-normalized jet uses the nominal scale $s$, so its block of order $k$ acquires the factor $(s/(s+b))^{k/2}$ relative to the normalized jet at the effective scale (\cref{prop-blur}). The graded retraction inherits the blur shift and is a complete invariant for the dilations of the nonzero jet; consequently every dilation-invariant feature of the raw jet factors through it. \Cref{tab-local} compares the three representations. The first two fail in different ways: the scale-normalized jet is invariant under dilations but is distorted by a blur, and the radial projection of the raw jet is not distorted but is not invariant under dilations. The graded retraction has both properties. The second of them is a shift and not an invariance to blur: the jet moves to a coarser scale, and the features pooled over positions and scales change with it.

\begin{table}[htbp]
\centering\small
\caption{Three local representations of the Gaussian jet. The last column refers to a Gaussian blur $I\mapsto g_b*I$ of the image without a change of contrast (\cref{prop-blur}).}\label{tab-local}
\begin{tabular}{@{}lll@{}}
\toprule
Local representation & \parbox[t]{0.27\textwidth}{\raggedright Invariant under dilation at corresponding scales} & \parbox[t]{0.27\textwidth}{\raggedright Gaussian blur acts by $s\mapsto s+b$ alone} \\
\midrule
Scale-normalized jet & yes & no \\
Radial projection of the raw jet & no, in general & yes \\
Graded retraction & yes & yes \\
\bottomrule
\end{tabular}
\end{table}

The retraction is singular at the zero jet, and there it keeps what other normalizations lose. Take the two jets $(tu_1,\pm t^2u_2)$, with a gradient $u_1$ and a Hessian form $u_2$ of norm one. As $t\to0$ both tend to zero. Their radial projections $(u_1,\pm tu_2)/\sqrt{1+t^2}$ come arbitrarily close to each other, while their retractions are the two fixed points $(u_1/\sqrt2,\pm u_2/2)$ for every $t$. \Cref{sec-separation} realizes this example by images: smooth images $I_t^\pm$ that converge uniformly to one image as $t\to0$, have these jets at one probe, and agree in their jets at all other probes, in their mass and in their intensity frame. A probe classifier built on the retraction separates the two families with the labels $\pm1$ for every $t$, while every radial probe classifier of \cref{def-probe} using the raw or the scale-normalized jet, and every classifier that is continuous in the image, has worst-case squared loss at least one on them (\cref{thm-local-separation,cor-continuous-cnn-separation}). The separation belongs to the ideal quotient. With a fixed positive regularization of the norm, which an implementation uses, the graded classifiers lose it on this family as well (\cref{prop-regularized-collapse}), so the result also measures what a regularization costs near the singular point.

The graded convolutional network of \cref{sec-networks} computes Gaussian jets on a geometric ladder of scales, retracts them, applies layers that are equivariant under rotations, built from complex-linear maps, products and gates on harmonic coordinates, and pools over positions and scales in a frame of the image that is covariant under dilations and translations. It differs from Lindeberg's Gaussian derivative networks in the normalization of the jet, which is by the jet itself and not by the scale; the scale enters only through a foreground gate, which uses the energy $s^{1/2}\rhoq(f)$, and through the pooling. For the ideal network on a ladder that doubles the scale $s$ at each step, the pooled features are invariant under the dilations by powers of two, when the sampling grids match and the pooling bins lie inside the ladder (\cref{prop-equiv}), and the network approximates every continuous function of the jet that is invariant under dilations and rotations, uniformly on compact sets (\cref{thm-universal}). Along the orbit of a jet under the grading action a ReLU network on the raw jet is eventually a polynomial in the orbit parameter, a network on the radial projection converges at either end to a function of a single derivative order, and the graded network is constant (\cref{thm-extrapolation}).

One experiment isolates the normalization. In our reimplementation of Lindeberg's network, trained on sharp digits, replacing the scale-normalized jet of the first layer by its retraction and changing nothing else, not even the number of parameters, raises the accuracy on MNIST digits blurred by a Gaussian of standard deviation two pixels from $91.45\%$ to $95.10\%$ (\cref{sub-exp-gdn}). \Cref{sec-exp} reports this experiment and the comparisons of three shallow networks that share the architecture and the training protocol and differ in the normalization of the jet and in the energy used by the foreground gate, on MNIST and on Fashion-MNIST, under zoom and under blur, including the conditions in which the graded network is the less accurate.

The jet has represented local image structure since Koenderink \cite{Koenderink1984, KoenderinkVanDoorn1987, Florack1992}, and blurring and scaling as operators on jet space were studied by Markussen, Pedersen and Loog \cite{MarkussenPedersenLoog2005, MarkussenPedersenLoog2008}. The classification of the local graded convolutions is a specialization of the distribution theory of homogeneous kernels \cite{Hormander1990}; equivariant differential operators for rotations and Euclidean motions were characterized by Jenner and Weiler \cite{JennerWeiler2022}. Complex features labelled by a rotation frequency and combined by products that add the frequencies are the charge-conserving networks of Yarotsky \cite{Yarotsky2022}, who proved their universality; the layers of the graded network are of this kind, with the derivative order as a second label. Scale-equivariant networks with learned filters \cite{Kanazawa2014, WorrallWelling2019, Sosnovik2020} also carry a ladder of scales. Exact invariants to an unknown Gaussian blur, and to wider families of convolutions, have been constructed from image moments and projection operators by Flusser and coauthors \cite{FlusserEtAl2016, KostkovaEtAl2020, FlusserEtAl2023}. The graded retraction is not such an invariant (\cref{sub-blur}); our concern is how a scale-invariant representation responds to blur.

\Cref{sec-jets} introduces graded jets, homogeneous norms and weighted spheres. \Cref{sec-conv} classifies the local graded convolutions and \cref{sec-invariants} the invariant local features, and \cref{sec-blur} treats blur and the passage between scales. \Cref{sec-networks} defines the graded networks and proves their invariance and universality, \cref{sec-separation} proves the separation at the zero jet, and \cref{sec-exp} reports the experiments. The appendix describes the discretization and its numerical verification. The code, the trained networks and the result files from which every table and figure is generated are deposited on Zenodo \cite{ShaskaCode2026}.

\section{Graded jets}\label{sec-jets}
In \cref{sec-jets,sec-conv,sec-invariants,sec-blur,sec-networks,sec-separation} images are functions on the plane, derivatives are exact Gaussian derivatives, and the homogeneous norm is unregularized except in \cref{prop-regularized-collapse}.

\subsection{Jets, dilations and the weighted retraction}\label{sub-ladder}\label{sub-retraction}
Derivatives of an image are taken after smoothing with the Gaussian $g_s(x)=(2\pi s)^{-1}\exp(-\abs{x}^2/2s)$, which is not homogeneous but covariant, $g_s*(\Dil_\lambda I)=\Dil_\lambda(g_{s/\lambda^2}*I)$, where $(\Dil_\lambda I)(x)=I(x/\lambda)$ is the magnification by $\lambda$. Write $L(x;s)=(g_s*I)(x)$, $L_\alpha=\partial^\alpha L$, and collect the derivatives of order $k$ into the binary form
\begin{equation}\label{eq-fk}
f_k(\xi,\eta;x,s)=\sum_{a+b=k}\binom{k}{a}L_{(a,b)}(x;s)\,\xi^a\eta^b\in V_k=\Sym^k(\R^2)^*,
\end{equation}
so that $L(x+h;s)=\sum_{k}f_k(h)/k!$ formally. The $N$-jet is $f=(f_0,\dots,f_N)\in V=\bigoplus_{k\le N}V_k$, with $\dim V_k=k+1$ and $\dim V=\frac{(N+1)(N+2)}{2}$; for $N=3$, $\dim V=10$, $f_1$ is the gradient, $f_2$ the Hessian form and $f_3$ a binary cubic. The grading action is $\mu\star f=(\mu^kf_k)_k$.

\begin{lemma}\label{lem-bundle}
For every $\lambda>0$, $x\in\R^2$ and $s>0$,
\begin{equation}\label{eq-dil}
f^{\Dil_\lambda I}(x;s)=\lambda^{-1}\star f^{I}(x/\lambda;\,s/\lambda^2).
\end{equation}
Consequently the normalized jet $\tilde f(x;s)=s^{1/2}\star f(x;s)$, that is $\tilde f_k=s^{k/2}f_k$, satisfies $\tilde f^{\Dil_\lambda I}(x;s)=\tilde f^I(x/\lambda;s/\lambda^2)$.
\end{lemma}
\begin{proof}
By the covariance of the Gaussian and the chain rule, $L^{\Dil_\lambda I}_\alpha(x;s)=\lambda^{-\abs\alpha}L^I_\alpha(x/\lambda;s/\lambda^2)$. For the second statement, $s^{1/2}\star\lambda^{-1}\star=(s/\lambda^2)^{1/2}\star$.
\end{proof}

The scale $s$ is thus a homogeneous function of degree two on the bundle of jets, and $\tilde f=\sqrt s\star f$ is the normalization of the jet by the homogeneous function $\sqrt s$ of degree one. This is the $\gamma$-normalization of \cite{Lindeberg1998} with $\gamma=1$; \cref{thm-forced} shows that the exponent is forced.

Let $\btheta=(\theta_1,\dots,\theta_N)$, $\theta_k>0$, be a vector of grades with the action $\mu\star_{\btheta}f=(\mu^{\theta_k}f_k)_k$ on $V^+=\bigoplus_{k=1}^NV_k$, and let $\norm{f_k}$ be a Euclidean norm on $V_k$ that is invariant under the rotations $f_k\mapsto f_k\circ\Rot^{-1}$, $\Rot\in\SO(2)$, of the plane. Define
\begin{equation}\label{eq-homnorm}
\rhot(f)=\Bigl(\sum_{k=1}^N\norm{f_k}^{2/\theta_k}\Bigr)^{1/2},\qquad
\pi_{\btheta}(f)_k=\frac{f_k}{\rhot(f)^{\theta_k}} .
\end{equation}
\begin{lemma}\label{lem-rho}
$\rhot$ is continuous on $V^+$, positive off $0$, $\SO(2)$-invariant and homogeneous of degree one, $\rhot(\mu\star_{\btheta}f)=\mu\rhot(f)$. The weighted sphere $S_{\btheta}=\{\rhot=1\}$ is compact, and $\pi_{\btheta}$ is a continuous retraction of $V^+\smallsetminus0$ onto $S_{\btheta}$ that is constant along the orbits of $\star_{\btheta}$. For $\btheta=(1,\dots,1)$, $\rhot$ is the Euclidean norm and $\pi_{\btheta}$ the radial projection.
\end{lemma}
\begin{proof}
$\norm{\mu^{\theta_k}f_k}^{2/\theta_k}=\mu^2\norm{f_k}^{2/\theta_k}$, and $\pi_{\btheta}(\mu\star_{\btheta}f)_k=\mu^{\theta_k}f_k/(\mu\rhot(f))^{\theta_k}$.
\end{proof}

This is the homogeneous norm of \cite{ShaskaGNN2025}, written with the exponents $2/\theta_k$ so that it is defined for real grades. The grades of the image jet are $\bq=(1,2,\dots,N)$, and $\pi_{\bq}$ is the quotient map of $V^+\smallsetminus0$ by the grading action. We write $\pi_{\mathbf1}$ for the radial projection. By \cref{lem-bundle}, $\rhoq(\tilde f)=s^{1/2}\rhoq(f)$ and $\pi_{\bq}(\tilde f)=\pi_{\bq}(f)$: the retraction of the normalized jet is the retraction of the raw jet. The scalar $\tilde\rho=\rhoq(\tilde f)$ is invariant under dilations at the corresponding scale; it is a scale-normalized energy of the jet of the kind used for scale selection \cite{Lindeberg1998}, and it is not invariant under contrast, $\rhoq(cf)^2=\sum_kc^{2/k}\norm{f_k}^{2/k}$.

\begin{lemma}\label{lem-regularity}
Let $N\ge2$. The retraction $\pi_{\bq}$ is smooth on the open set where every block $f_k$ is nonzero and locally H\"older continuous of exponent $\min(1,2/N)$ on $V^+\smallsetminus0$, and for $N\ge3$ this exponent cannot be improved. It has no continuous extension to $0$.
\end{lemma}
\begin{proof}
The function $f_k\mapsto\norm{f_k}^{2/k}=(\norm{f_k}^2)^{1/k}$ is smooth where $f_k\ne0$ and locally H\"older continuous of exponent $\min(1,2/k)$ on $V_k$. Hence $\rhoq^2$ is locally H\"older of exponent $\min(1,2/N)$, and so is $\rhoq^{-k}=(\rhoq^2)^{-k/2}$ on every compact subset of $V^+\smallsetminus0$, where $\rhoq$ is bounded below; the components $f_k\rhoq^{-k}$ of $\pi_{\bq}$ inherit both properties. For $N\ge3$ let $u_1\in V_1$ and $v\in V_N$ be unit vectors and $f_t=(u_1,0,\dots,0,tv)$. Then $\rhoq(f_t)^2=1+\abs t^{2/N}$ and $\norm{\pi_{\bq}(f_t)_1-\pi_{\bq}(f_0)_1}=1-(1+\abs t^{2/N})^{-1/2}\sim\frac12\abs t^{2/N}$ as $t\to0$. Finally, for unit vectors $u_1\in V_1$ and $u_2\in V_2$ the curves $t\mapsto(tu_1,\pm t^2u_2,0,\dots,0)$ tend to $0$ as $t\to0$, while $\rhoq^2=2t^2$ along both and their retractions are the constant points $(u_1/\sqrt2,\pm u_2/2,0,\dots,0)$.
\end{proof}

\subsection{Rotations and contrast}\label{sub-forms}\label{sub-group}
Two further transformations change an image without changing what a recognizer should respond to: rotations, and the contrast changes $I\mapsto cI+d$ with $c>0$, which act by $f_0\mapsto cf_0+d$ and $f_k\mapsto cf_k$ for $k\ge1$. On $V^+$ the torus $T=\Gm\times\Gm$ of dilations and contrasts acts by $(\mu,c)\star f_k=\mu^kc\,f_k$, and $\SO(2)$ acts on each $V_k$ and commutes with $T$. Let $G=T\times\SO(2)$.

Rotations $\Rot\in\SO(2)$ act on images by $(\Rot\cdot I)(x)=I(\Rot^{-1}x)$ and on the jet by $f_k\mapsto f_k\circ\Rot^{-1}$, the restriction to $\SO(2)$ of the action of $\GL_2$ on binary forms; they commute with dilations. Under $\SO(2)$,
\begin{equation}\label{eq-harmonic}
V_k=\Hfreq_k\oplus(\xi^2+\eta^2)\Hfreq_{k-2}\oplus\cdots,
\end{equation}
where $\Hfreq_j$, the harmonic forms of degree $j$, is the irreducible representation of frequency $j$, of dimension $2$ for $j\ge1$ and $1$ for $j=0$; this is the decomposition behind steerable filters \cite{FreemanAdelson1991, WeilerCesa2019}. In the complex coordinate $z=\xi+i\eta$ write $f_k=\sum_{a+b=k}c_{k,b-a}\,z^a\bar z^b$, with $c_{k,-j}=\bar c_{k,j}$. The rotation by $\phi$ sends $z\mapsto e^{-i\phi}z$ and $\bar z\mapsto e^{i\phi}\bar z$, so
\begin{equation}\label{eq-freq}
c_{k,j}\mapsto e^{ij\phi}c_{k,j},\qquad j=k,k-2,\dots,\ j\ge0.
\end{equation}
The \emph{harmonic coordinates} $c_{k,j}$, $j\ge0$, are complex numbers, real for $j=0$, on which $\SO(2)$ acts diagonally by the frequency $j$ and the grading action by $\mu^k$. For $N=3$ there are five of them, of order and frequency $(1,1)$, $(2,0)$, $(2,2)$, $(3,1)$ and $(3,3)$. For the $2$-jet they are, up to factors, the classical complex coordinates
\[
z_1=L_x+iL_y,\qquad h_0=\tfrac12(L_{xx}+L_{yy}),\qquad h_2=\tfrac12(L_{xx}-L_{yy})+iL_{xy},
\]
which are twice the coefficient of $\bar z$ in $f_1$, the coefficient of $z\bar z$ in $f_2$ and twice the coefficient of $\bar z^2$ in $f_2$. The classical differential invariants of second order are polynomials of frequency zero in them: $\abs{\nabla L}^2=z_1\bar z_1$, $\Lap L=2h_0$, $\det\Hess L=h_0^2-h_2\bar h_2$ and $\nabla L^{\!\top}\Hess L\,\nabla L=h_0z_1\bar z_1+\operatorname{Re}(\bar h_2z_1^2)$. From here on $\norm{f_k}$ is the Euclidean norm of the harmonic coordinates of $f_k$.

\begin{proposition}\label{prop-dim}
For $N\ge2$ the generic $G$-orbit on $V^+$ has dimension $3$, so the space of $N$-jets modulo scale, rotation and contrast has dimension $\frac{(N+1)(N+2)}{2}-4$: $2$ for $N=2$, $6$ for $N=3$ and $11$ for $N=4$.
\end{proposition}
\begin{proof}
On the open dense set where $f_1\ne0$ and the trace component of $f_2$ is nonzero, let $a$, $c$ and $b$ be infinitesimal dilation, contrast and rotation parameters that fix $f$. In grade one, $(a+c)f_1+bJf_1=0$, where $J$ is the rotation by $\pi/2$; since $Jf_1\perp f_1$, $a+c=0$ and $b=0$. The trace component of grade two, on which rotations act trivially, gives $2a+c=0$, hence $a=c=0$. The infinitesimal stabilizer is zero, so the generic orbit has dimension three.
\end{proof}

The quotient of $V^+\smallsetminus0$ by the positive dilations is the weighted sphere $S_{\bq}$, of dimension $\dim V^+-1$, which is $8$ for $N=3$. We use positive dilations throughout.

Which of the three factors of $G$ is a nuisance depends on the task. Scale is one throughout this paper. Contrast is removed at the input by $\hat I=(I-\min I)/(\max I-\min I)$, which commutes with dilations and rotations; a negative $c$, which gives $1-\hat I$, is not removed, and constant images are mapped to zero. Rotation is a nuisance for textures but not for upright digits, a $6$ being a rotated $9$; the networks of \cref{sec-networks} are equivariant under rotations, and their readout decides whether the orientation is kept.

\section{Graded convolutions}\label{sec-conv}
Let $V=\bigoplus_{k=0}^NV_k$ be a graded real vector space. A \emph{graded feature field} is a map $F=(F_0,\dots,F_N):\R^2\to V$. Translations act by $(T_aF)(x)=F(x-a)$, and dilations by
\begin{equation}\label{eq-dilfield}
(\Dil_\lambda F)_k(x)=\lambda^{-k}F_k(x/\lambda),\qquad \lambda>0,
\end{equation}
the change of variable of a magnification by $\lambda$ combined with the grading action $\lambda^{-1}\star$ on the fibre \cite{ShaskaGNN2025}. An image $I:\R^2\to\R$ is a field of grade $0$, on which \eqref{eq-dilfield} is the magnification $(\Dil_\lambda I)(x)=I(x/\lambda)$.

\begin{definition}\label{def-gradedconv}
A \emph{graded convolution} from grade $k$ to grade $k'$ is a continuous linear operator $L$ from fields of grade $k$ to fields of grade $k'$ that commutes with all translations and all dilations, $LT_a=T_aL$ and $L\Dil_\lambda=\Dil_\lambda L$. It is \emph{local} if $\supp LF\subseteq\supp F$ for every $F$.
\end{definition}

\begin{proposition}\label{prop-gradedconv}
Let $L:C_c^\infty(\R^2)\to\mathcal D'(\R^2)$ be a graded convolution from grade $k$ to grade $k'$. Then $LF=K*F$ for a distribution $K$ that is homogeneous of degree $k-k'-2$. If $L$ is nonzero and local, then $k'\ge k$ and
\[
L=\sum_{\abs{\alpha}=k'-k}c_\alpha\,\partial^\alpha ,
\]
a homogeneous differential operator of order $m=k'-k$ with constant coefficients. Conversely, every such operator is a local graded convolution from grade $k$ to grade $k+m$.
\end{proposition}
\begin{proof}
A continuous linear operator on $C_c^\infty(\R^2)$ that commutes with translations is the convolution with a distribution $K$ \cite{Hormander1990}. For the dilations compute, with $y=\lambda u$,
\begin{align*}
(K*\Dil_\lambda F)(x)&=\lambda^{-k}\!\int\! K(x-y)F(y/\lambda)\,dy=\lambda^{2-k}\!\int\! K(x-\lambda u)F(u)\,du,\\
(\Dil_\lambda(K*F))(x)&=\lambda^{-k'}\!\int\! K(x/\lambda-u)F(u)\,du .
\end{align*}
Equality for all $F$ gives, with $z=x/\lambda-u$, $K(\lambda z)=\lambda^{k-k'-2}K(z)$ in the sense of distributions, so $K$ is homogeneous of degree $k-k'-2$. Locality gives $\supp K\subseteq\{0\}$, and a distribution supported at a point is a finite sum $\sum_\alpha c_\alpha\partial^\alpha\delta$ \cite{Hormander1990}. Since $\partial^\alpha\delta$ is homogeneous of degree $-2-\abs{\alpha}$ and distributions of distinct degrees of homogeneity are linearly independent, $\abs{\alpha}=k'-k$ whenever $c_\alpha\ne0$; in particular $k'\ge k$ when $L\ne0$. The converse is the chain rule, $\partial^\alpha(F(\cdot/\lambda))=\lambda^{-\abs{\alpha}}(\partial^\alpha F)(\cdot/\lambda)$.
\end{proof}

\begin{corollary}\label{cor-jet}
The $N$-jet map $I\mapsto(\partial^\alpha I)_{\abs{\alpha}\le N}$, with $\partial^\alpha I$ placed in grade $\abs{\alpha}$, is a local graded convolution from grade $0$ to $\bigoplus_{k\le N}V_k$, and every local graded convolution from grade $0$ of order at most $N$ factors through it by a grade-preserving linear map.
\end{corollary}

The map $I\mapsto f(\cdot;s)$ of \cref{sub-ladder} is this graded convolution applied to $g_s*I$.

\begin{remark}\label{rem-riesz}
Without locality there are graded convolutions of grade shift zero other than the multiples of the identity. The distributions on $\R^2$ that are homogeneous of degree $-2$ are $c\,\delta+\mathrm{p.v.}\,\Omega(z/\abs z)/\abs z^2$ with $\Omega$ of mean zero on the circle \cite{Hormander1990}, and the singular integrals among them are nonlocal graded convolutions of grade shift zero. The Riesz transforms, with kernels $z_j/\abs{z}^3$, which underlie the monogenic signal \cite{FelsbergSommer2001}, are of this kind. They could be added to the first layer, where their two components transform as a vector under rotations; we do not use them.
\end{remark}

\section{Invariants of the dilation action}\label{sec-invariants}
Let $\bq=(1,\dots,N)$. By \cref{lem-bundle} the dilation $\Dil_\lambda$ carries the jet $f$ of $I$ at $(x;s)$ to the jet $\lambda^{-1}\star f$ of $\Dil_\lambda I$ at $(\lambda x;\lambda^2s)$, so it acts on pairs of a jet and a scale by
\begin{equation}\label{eq-bundleaction}
\lambda\cdot(f,s)=(\lambda^{-1}\star f,\ \lambda^2s),\qquad\lambda>0.
\end{equation}
A \emph{local feature} is a map $\Phi:V^+\times(0,\infty)\to Y$ into a set $Y$. It is \emph{scale-invariant} if $\Phi(\lambda\cdot(f,s))=\Phi(f,s)$ for all $\lambda$, $f$ and $s$. Every element of $V^+$ is the jet at $(0;s)$ of an image in $C_c^\infty(\R^2)$: the map $I\mapsto f^I(0;s)$ is linear, and a linear form vanishing on its image would be a nonzero combination of derivatives of $g_s$ orthogonal to every test function, which is impossible since these derivatives are linearly independent. Scale invariance of $\Phi$ is therefore equivalent to $\Phi\bigl(f^{\Dil_\lambda I}(\lambda x;\lambda^2s),\lambda^2s\bigr)=\Phi\bigl(f^I(x;s),s\bigr)$ for all images, positions and scales. A feature that does not use the scale is a map $\Psi:V^+\smallsetminus0\to Y$; it is scale-invariant if $\Psi(\lambda\star f)=\Psi(f)$ for all $\lambda>0$.

For a vector $\btheta$ of positive weights let $\star_{\btheta}$, $\rhot$ and $\pi_{\btheta}$ be as in \cref{sub-retraction}, and let
\[
\nu_{\btheta}(f,s)=s^{1/2}\star_{\btheta}f=(s^{\theta_k/2}f_k)_k,\qquad \nu=\nu_{\bq},
\]
so that $\nu(f,s)=\tilde f$ is the normalized jet of \cref{lem-bundle}. Lindeberg's $\gamma$-normalized derivatives \cite{Lindeberg1998} form $\nu_{\gamma\bq}$.

\begin{theorem}\label{thm-forced}
\begin{enumerate}[label=(\roman*)]
\item A local feature $\Phi$ is scale-invariant if and only if $\Phi=\Phi_0\circ\nu$ for a map $\Phi_0:V^+\to Y$, and then $\Phi_0=\Phi(\cdot,1)$. The map $\nu$ is scale-invariant and surjective, and its fibres are the orbits of \eqref{eq-bundleaction}.
\item A feature $\Psi$ that does not use the scale is scale-invariant if and only if $\Psi=\Psi_0\circ\pi_{\bq}$ for a map $\Psi_0:S_{\bq}\to Y$, and then $\Psi_0=\Psi|_{S_{\bq}}$.
\item $\pi_{\bq}\circ\nu(f,s)=\pi_{\bq}(f)$ and $\rhoq(\nu(f,s))=s^{1/2}\rhoq(f)$, and $(f,s)\mapsto\bigl(\pi_{\bq}(f),s^{1/2}\rhoq(f)\bigr)$ maps the orbit space of $(V^+\smallsetminus0)\times(0,\infty)$ bijectively onto $S_{\bq}\times(0,\infty)$.
\item For all weights $\btheta$, $\nu_{\btheta}(\lambda\cdot(f,s))=\lambda^{\btheta-\bq}\star\nu_{\btheta}(f,s)$, where $\lambda^{\btheta-\bq}\star$ multiplies $V_k$ by $\lambda^{\theta_k-k}$. Hence $\nu_{\btheta}$ is scale-invariant if and only if $\btheta=\bq$, and $\Phi_0\circ\nu_{\btheta}$ is scale-invariant if and only if $\Phi_0$ is constant on the orbits of the one-parameter group $\{\lambda^{\btheta-\bq}\star\}_{\lambda>0}$, which is nontrivial for $\btheta\ne\bq$.
\item The retraction $\pi_{\btheta}$ is scale-invariant if and only if $\btheta=c\bq$ for some $c>0$. For $N\ge2$ and every other $\btheta$, $\Psi_0\circ\pi_{\btheta}$ is scale-invariant if and only if $\Psi_0$ is constant along the curves $\lambda\mapsto\pi_{\btheta}(\lambda\star u)$, $u\in S_{\btheta}$, which are not all constant; in particular $\Psi_0$ is not injective.
\end{enumerate}
\end{theorem}
\begin{proof}
(i) $\nu(\lambda\cdot(f,s))=(\lambda^2s)^{1/2}\star\lambda^{-1}\star f=s^{1/2}\star f$, since $\star$ is an action, and $\nu(f,1)=f$, so $\nu$ is surjective. If $\nu(f,s)=\nu(f',s')$, then $(f',s')=\lambda\cdot(f,s)$ with $\lambda=(s'/s)^{1/2}$, so the fibres of $\nu$ are the orbits. An invariant $\Phi$ therefore satisfies $\Phi(f,s)=\Phi\bigl(s^{-1/2}\cdot(f,s)\bigr)=\Phi(\nu(f,s),1)$, and every $\Phi_0\circ\nu$ is invariant. (ii) $\pi_{\bq}$ is constant on the orbits of $\star$ and the identity on $S_{\bq}$, and $f=\rhoq(f)\star\pi_{\bq}(f)$ (\cref{lem-rho}); so $\Psi(f)=\Psi(\pi_{\bq}(f))$ for an invariant $\Psi$, and the converse is clear. (iii) The two identities are those of \cref{lem-rho} with $\mu=s^{1/2}$. The map is invariant by (i), and its value $(u,r)$ determines $\nu(f,s)=(s^{1/2}\rhoq(f))\star\pi_{\bq}(f)=r\star u$, hence the orbit; every $(u,r)$ is attained at $(r\star u,1)$. (iv) $(\lambda^2s)^{\theta_k/2}\lambda^{-k}=\lambda^{\theta_k-k}s^{\theta_k/2}$; the rest follows as in (i), since $\nu_{\btheta}(\cdot,1)$ is the identity. (v) The fibres of $\pi_{\btheta}$ are the orbits of $\star_{\btheta}$, so $\pi_{\btheta}$ is scale-invariant if and only if for all $\lambda$ and $f$ there is $\mu>0$ with $\lambda\star f=\mu\star_{\btheta}f$. For $\btheta=c\bq$ take $\mu=\lambda^{1/c}$. Conversely, for $f$ with all blocks nonzero, $\lambda f_1=\mu^{\theta_1}f_1$ and $\lambda^kf_k=\mu^{\theta_k}f_k$ give $\mu=\lambda^{1/\theta_1}$ and $\theta_k=k\theta_1$ for every $k$. Since the actions $\star$ and $\star_{\btheta}$ commute and $f=\rhot(f)\star_{\btheta}\pi_{\btheta}(f)$, one has $\pi_{\btheta}(\lambda\star f)=\pi_{\btheta}(\lambda\star\pi_{\btheta}(f))$, which gives the last statement; for $\btheta$ not proportional to $\bq$ and $u$ with all blocks nonzero the curve is not constant, by the computation just made.
\end{proof}

Case (i) is the scale normalization on which the Gaussian derivative networks \cite{Lindeberg2022} are built, and case (ii) is the graded retraction, which needs no scale. A normalization of the raw jet with equal grades, or with grades not proportional to $\bq$, is excluded by (iv) and (v). The theorem does not require the weights $\bq$ in the first normalization of a network: since $\pi_{\bq}(\nu_{\gamma\bq}(f,s))=\pi_{\bq}(f)$ for every $\gamma>0$, an invariant feature may follow a $\gamma$-normalization that is not itself invariant, and by (iv) and (ii) it then factors through $\pi_{\bq}$. The two factorizations differ in content: the bundle quotient keeps the invariant $s^{1/2}\rhoq(f)$ of part (iii), which the fibre quotient discards, and which the networks of \cref{sec-networks} use only in a foreground gate. Under blur the two realizations behave differently.

\section{Blur and the scale direction}\label{sec-blur}
\subsection{Unknown blur}\label{sub-blur}
An image may arrive already smoothed: by defocus, by a reduction of resolution, or by optics whose blur is fixed in pixels while the object is seen smaller. We model this by $I'=g_b*I$ with an unknown $b>0$. Since $g_s*g_b=g_{s+b}$, the jet of $I'$ at the scale $s$ is the jet of $I$ at the scale $s+b$. For an image $I\ge0$ of finite positive mass and finite second moment, let $\bar x_I$ and $\hat\sigma_I$ be the mean and the root-mean-square radius of the probability density $I/\int I$, so that $\hat\sigma_I^2=\int\abs{x-\bar x_I}^2I(x)\,dx/\int I$; we call $(\bar x_I,\hat\sigma_I)$ the \emph{intensity frame} of $I$.

\begin{proposition}\label{prop-blur}
Let $b>0$, $I'=g_b*I$, $\mu=\bigl(s/(s+b)\bigr)^{1/2}\in(0,1)$, and let $u=\nu\bigl(f^I(x;s+b),s+b\bigr)$ be the normalized jet of $I$ at the scale $s+b$. For every $x$, every $s>0$ and every $c>0$:
\begin{enumerate}[label=(\roman*)]
\item $f^{cI'}(x;s)=c\,f^I(x;s+b)$, and for $J=c\,(g_b*\Dil_\lambda I)$, $f^J(\lambda x;s)=c\,\lambda^{-1}\star f^I\bigl(x;(s+b)/\lambda^2\bigr)$;
\item $\pi_{\bq}\bigl(f^{I'}(x;s)\bigr)=\pi_{\bq}\bigl(f^I(x;s+b)\bigr)$, whereas $\nu\bigl(f^{I'}(x;s),s\bigr)=\mu\star u$, and $\pi_{\mathbf1}(\mu\star u)\ne\pi_{\mathbf1}(u)$ unless all nonzero blocks of $u$ have the same order;
\item $\pi_{\bq}\bigl(f^{cI'}(x;s)\bigr)=\pi_{\bq}(c\,u)$ and $\pi_{\mathbf1}\bigl(\nu(f^{cI'}(x;s),s)\bigr)=\pi_{\mathbf1}(\mu\star u)$;
\item $s^{1/2}\rhoq\bigl(f^{cI'}(x;s)\bigr)=\mu\,\rhoq(c\,u)$ and $\norm{\nu(f^{cI'}(x;s),s)}=c\bigl(\sum_k\mu^{2k}\norm{u_k}^2\bigr)^{1/2}$;
\item if $I\ge0$ has finite positive mass and finite second moment, then $\bar x_{J}=\lambda\bar x_I$ and $\hat\sigma_{J}^2=\lambda^2\hat\sigma_I^2+2b$.
\end{enumerate}
In (ii)--(iv) the same holds for $J$, with $(s+b)/\lambda^2$ in place of $s+b$ and the same $\mu$.
\end{proposition}
\begin{proof}
(i) $g_s*g_b=g_{s+b}$, linearity, and \cref{lem-bundle}. (ii) and (iii) $s^{1/2}\star f=\mu\star\bigl((s+b)^{1/2}\star f\bigr)$, since $\star$ is an action and commutes with scalar multiplication; $\pi_{\bq}$ is constant on the orbits of $\star$ (\cref{lem-rho}) and $\pi_{\mathbf1}$ on the rays $u\mapsto cu$. If $\pi_{\mathbf1}(\mu\star u)=\pi_{\mathbf1}(u)$ with $\mu\ne1$, then $\mu^ku_k=a\,u_k$ for one $a>0$ and every nonzero block, which forces the orders of the nonzero blocks to coincide. (iv) The homogeneity of $\rhoq$ under $\star$, and the definition of the Euclidean norm. (v) Normalize the intensity to a probability density. Convolution with $g_b$ adds an independent centred Gaussian vector of covariance $b\,\mathrm{Id}$, which keeps the mean and adds its trace $2b$ to the squared radius; a dilation multiplies the mean by $\lambda$ and the squared radius by $\lambda^2$, and $c$ does not change the density. For $J$, apply (i) and \cref{lem-bundle}: $s^{1/2}\star\lambda^{-1}\star f=\mu\star\bigl((s+b)/\lambda^2\bigr)^{1/2}\star f$.
\end{proof}

Blur acts on the retracted jet and on the normalized jet alike by the change of scale $s\mapsto s+b$, and on the normalized jet in addition by the grading factor $\mu\star$, with $\mu^2=s/(s+b)$ smallest at the finest scales. The change of scale is shared by every function of the raw jet, the radial projection $\pi_{\mathbf1}(f)$ included, since by (i) the raw jet of $I^{\prime}$ at the scale $s$ is that of $I$ at the scale $s+b$; the grading factor comes from the normalization by the nominal scale. What distinguishes $\pi_{\bq}$ among the functions of the raw jet is that it is also a complete invariant of the dilation action (\cref{thm-forced}(ii)). Without a contrast factor the retracted jet of the blurred image is that of the sharp image at a coarser scale, of the kind a network evaluated on a ladder of scales encounters in training. The radial projection of the normalized jet does not remove the grading factor, so the normalized jet of the blurred image differs from that of the same image at the scale $s+b$ by $\mu\star$, most at the finest scales. By (iii), a contrast factor behaves the other way: the retraction keeps it and the radial projection removes it. Normalizing by $s+b$ in place of $s$ would correct the scale normalization, but only for a known $b$; the retraction needs neither $s$ nor $b$. Lindeberg's networks normalize the derivatives of each layer by the nominal scale of that layer, so (ii) applies to their first layer, while their later layers smooth further, at scales that grow with depth.

\Cref{prop-blur} concerns features and is not an invariance statement. First, the retraction is not invariant under blur at a fixed position and scale: for $I(x)=e^{-\abs x^2/2}$ and $x_s=(\sqrt{1+s},0)$, the second derivative of $g_s*I$ in $x_1$ vanishes at $x_s$ while the first does not, and after a blur of variance $b$ it equals $-b(1+s+b)^{-3}\exp\bigl(-\frac{1+s}{2(1+s+b)}\bigr)\ne0$. The grading action multiplies the whole second block by one positive factor and cannot make a vanishing coordinate nonzero, so $\pi_{\bq}\bigl(f^{I'}(x_s;s)\bigr)\ne\pi_{\bq}\bigl(f^I(x_s;s)\bigr)$. Second, $s\mapsto s+b$ is an additive change, not a shift of a geometric ladder of scales, and by (iv) and (v) the scale-normalized energies and the intensity frame change as well, so a feature pooled over positions and scales need not be preserved. Third, a contrast normalization after the blur introduces a factor $c_b=1/\sup(g_b*I)$, which by (iii) the retraction keeps. How the robustness of the features carries over to predictions is measured in \cref{sub-exp-blur}.

\subsection{Zoom between sampled scales}\label{sub-between}
\begin{lemma}\label{lem-octave}
Let $\Lap:V_{k+2}\to V_k$ be the contraction $\Lap_k=\frac{1}{(k+1)(k+2)}(\partial_\xi^2+\partial_\eta^2)$, acting componentwise on sequences $(f_k)_{k\ge0}$ of forms, and let $E$ be the grading operator, $Ef_k=kf_k$. Let $I$ be a bounded measurable image and $f=(f_k)_{k\ge0}$ the full jet of $L(\cdot;s)=g_s*I$.
\begin{enumerate}[label=(\roman*)]
\item $\partial_s f_k=\tfrac12\Lap_kf_{k+2}$ for $s>0$, and $[E,\Lap]=-2\Lap$.
\item For $0<s'<2s$ and $\mu=(s'/s)^{1/2}$ the series
\[
\tilde f_k(x;s')=\mu^k\sum_{j\ge0}\frac1{j!}\Bigl(\frac{\mu^2-1}{2}\Bigr)^j\Lap^j\tilde f_{k+2j}(x;s)
\]
converges for every $k$; we write it as
\begin{equation}\label{eq-octave}
\tilde f(x;s')=\mu\star\exp\!\Bigl(\tfrac{\mu^2-1}{2}\Lap\Bigr)\tilde f(x;s).
\end{equation}
If $I$ is a polynomial, the sums are finite and \eqref{eq-octave} holds for every $s'>0$.
\end{enumerate}
\end{lemma}
\begin{proof}
(i) With \eqref{eq-fk} and the heat equation $\partial_sL=\frac12(L_{xx}+L_{yy})$, the coefficient of $\binom ka\xi^a\eta^b$ in $\partial_sf_k$ is $\frac12(L_{(a+2,b)}+L_{(a,b+2)})$, and $\frac1{(k+1)(k+2)}(\partial_\xi^2+\partial_\eta^2)$ applied to \eqref{eq-fk} with $k+2$ in place of $k$ gives $\sum_{a+b=k}\binom ka(L_{(a+2,b)}+L_{(a,b+2)})\xi^a\eta^b$, as the identity $\binom{k+2}{a+2}(a+2)(a+1)=(k+2)(k+1)\binom ka$ shows. Since $\Lap$ lowers the degree by two, $E\Lap v=(k-2)\Lap v=\Lap Ev-2\Lap v$. (ii) By (i), $\partial_s^jf_k=2^{-j}\Lap^jf_{k+2j}$. For $\operatorname{Re}z>0$ the kernel $g_z(x)=(2\pi z)^{-1}\exp(-\abs x^2/2z)$ and its derivatives are integrable, so for bounded $I$ the function $z\mapsto(\partial^\alpha g_z*I)(x)$ is holomorphic on $\operatorname{Re}z>0$, and its Taylor series at $s$ converges to it on the disc $\abs{z-s}<s$. For $0<s'<2s$ this gives $f_k(s')=\sum_j\frac{(s'-s)^j}{2^jj!}\Lap^jf_{k+2j}(s)$, and then
\[
\tilde f_k(s')=s'^{k/2}\sum_{j}\frac{(s'-s)^j}{2^jj!}\Lap^jf_{k+2j}(s)
=\Bigl(\frac{s'}{s}\Bigr)^{k/2}\sum_j\frac{1}{j!}\Bigl(\frac{s'-s}{2s}\Bigr)^j\Lap^j\bigl(s^{(k+2j)/2}f_{k+2j}(s)\bigr),
\]
which is (ii) with $(s'-s)/s=\mu^2-1$. For a polynomial image $L(x;s)$ is a polynomial in $s$ and $\Lap^jf_{k+2j}=0$ for large $j$.
\end{proof}

The component of order $k$ at $s'$ involves all the orders $k+2j$ at $s$, so the $N$-jet at $s'$ is not a function of the $N$-jet at $s$. In \eqref{eq-octave} the factor $\mu\star$ is the grading action, and the factor $\exp(\frac{\mu^2-1}{2}\Lap)$ is triangular with respect to the derivative order, with the identity on the diagonal: it adds to each order a combination of the orders above it. On the finite jets of polynomial images it is unipotent; on the full jet of a general image the series in $\Lap$ does not terminate. Beyond $s'=2s$ the series need not converge: for $I(x)=e^{-\abs x^2/2a}$ one has $L(0;s)=a/(a+s)$, whose Taylor series at $s$ has radius $s+a$ and diverges for $s'>2s+a$, arbitrarily close to $2s$ as $a\to0$.

Let the jet be computed on a geometric ladder of scales $s_i=s_0\kappa^i$ with $1<\kappa\le2$, as in the networks of \cref{sec-networks}. Under a dilation by a factor $\lambda$ that is not a power of $\kappa^{1/2}$, the jet of $\Dil_\lambda I$ at $(\lambda x;s_j)$ is, by \cref{lem-bundle}, $\lambda^{-1}\star f^I(x;s_j/\lambda^2)$, the jet of $I$ at a scale strictly between two rungs. Comparing it with the jet of $I$ at the rung $s_i$ below is the situation of \cref{lem-octave} with $s=s_i$ and $s'=s_j/\lambda^2<2s_i$.

\begin{proposition}\label{prop-octave}
Let $0<s'<2s$, $\mu=(s'/s)^{1/2}$, and let
\[
R_k(s)=\sum_{j\ge1}\frac1{j!}\left(\frac{\mu^2-1}{2}\right)^j\Lap^j\tilde f_{k+2j}(s)
\]
be the heat correction of \cref{lem-octave}. For the jets of positive order at most $N$, whenever the jets involved are nonzero,
\[
\pi_{\bq}(\tilde f(s'))=\pi_{\bq}(\tilde f(s)+R(s)),\qquad
\pi_{\mathbf1}(\tilde f(s'))=\pi_{\mathbf1}(\mu\star(\tilde f(s)+R(s))).
\]
Let $h=(\tilde f_k(s))_{k=1}^N$, $a=(\Lap_k\tilde f_{k+2}(s))_{k=1}^N$, $\delta=\mu^2-1$ and $(Eh)_k=kh_k$. If every block of $h$ is nonzero, then, as $\delta\to0$,
\begin{align*}
\pi_{\bq}(\tilde f(s'))&=\pi_{\bq}(h)+\tfrac\delta2\,d\pi_{\bq}(h)[a]+O(\delta^2),\\
\pi_{\mathbf1}(\tilde f(s'))&=\pi_{\mathbf1}(h)+\tfrac\delta2\,d\pi_{\mathbf1}(h)[Eh+a]+O(\delta^2).
\end{align*}
\end{proposition}
\begin{proof}
The identities follow from \cref{lem-octave} and \cref{lem-rho}. The convergent heat expansion gives $h+\delta a/2+O(\delta^2)$ once the grading factor is removed. Since $\mu^k=(1+\delta)^{k/2}=1+k\delta/2+O(\delta^2)$, keeping that factor gives $h+\delta(Eh+a)/2+O(\delta^2)$. The two retractions are smooth at a point with nonzero blocks (\cref{lem-regularity}), and a Taylor expansion proves the formulas. The constants depend on the image, the position and the scale.
\end{proof}

To first order the retraction removes the grading term $Eh$ and keeps only the heat correction $a$, while the radial projection keeps both. The formulas do not by themselves order the sizes of the two residuals. The radial projection quotients the normalized jet by scalar multiplication, which is the contrast factor of \cref{sub-group}, and the retraction quotients it by the grading action; quotienting by both, that is by the torus $T$, would remove one more dimension (\cref{prop-dim}). Between two scales the normalized jet changes by the grading factor and the heat correction, not by a scalar factor (\cref{lem-octave}); with contrast removed at the input, the scalar factor that a radial normalization removes is the local intensity of the image, not an effect of the zoom. Which realization is closer between rungs depends on the size of the heat correction. For digit strokes, whose width is comparable to the finer rungs, it is not negligible (\cref{sub-verification}), so between rungs neither realization has a guaranteed advantage.

\section{Graded networks}\label{sec-networks}

\subsection{The network}\label{sub-networks}\label{sub-readout}\label{sub-model}\label{sub-invariance}
Fix an order $N$, grades $\btheta$ and a geometric ladder of scales $s_i=s_0\kappa^i$, $i=0,\dots,n-1$, with $\kappa>1$. A \emph{graded network with grades $\btheta$} computes, at every position $x$ of a sampling grid and every scale $s_i$, the jet $f(x;s_i)$ of \eqref{eq-fk}, retracts it to $\pi_{\btheta}(f(x;s_i))$, and applies to the retracted jet a pointwise map that is equivariant under rotations. The weights are shared across positions and scales.

The pointwise map acts on complex \emph{slots}. A slot has a type $(w,j)$, where $w$ is a formal weight, the total derivative order, and $j\ge0$ is a frequency; a rotation by $\phi$ multiplies a slot of frequency $j$ by $e^{ij\phi}$, and each type carries $m$ channels. The harmonic coordinates $c_{k,j}$ of the retracted jet are the input slots, of types $(k,j)$. There are three kinds of layers.
\begin{enumerate}[label=(\alph*)]
\item \emph{Complex-linear maps}, applied type by type with the same weights at every position and scale. For $j\ge1$ they are, by Schur's lemma, the general $\SO(2)$-equivariant real-linear maps between slots of frequency $j$; for $j=0$, where $\SO(2)$ acts trivially, they form a proper subset of the real-linear maps, which suffices for what follows. As convolutions with a $1\times1$ kernel and grade shift zero they commute with translations and, by \cref{prop-gradedconv}, with dilations; they are the graded linear maps of degree zero of \cite{ShaskaGNN2025}.
\item \emph{Products}. For slots $z$ of type $(w,j)$ and $z'$ of type $(w',j')$ the products $zz'$ and $z\bar z'$ have the types $(w+w',j+j')$ and $(w+w',\abs{j-j'})$, the conjugate being stored when $j<j'$. In the harmonic basis their complex multiples are the $\SO(2)$-equivariant bilinear maps into slots of positive frequency, and the real and imaginary parts of the products of frequency zero span, over $\R$, the invariant bilinear forms. They realize the graded tensor products of \cite{ShaskaGNN2025} compatibly with rotations, and the transvectants of classical invariant theory, which are the $\SL_2$-equivariant bilinear maps on binary forms \cite{Olver99}, are linear combinations of them; for instance $(f_2,f_2)^{(2)}=2\det\Hess L$. A product layer forms the channelwise products of its input slots with the input jet slots for all pairs of types with $w+w'\le w_{\max}$, appends its input slots, and maps each type to $m$ channels by a layer of kind (a).
\item \emph{Gates} $z\mapsto\sigma(\alpha(\abs z-\beta))\,z$, with $\sigma$ the logistic function, $\alpha$ fixed and $\beta$ learnable per channel; they are equivariant because $\abs z$ is invariant.
\end{enumerate}
After the retraction every slot is invariant under the grading action, so the gates need no normalization, and no other nonlinearity is used. The slots of the last layer are read as real numbers. The \emph{full readout} takes the real and imaginary parts of every slot, which transform under rotations and keep the orientation; the \emph{invariant readout} takes $\abs z$ for every slot and the real and imaginary parts of the slots of frequency zero. A learnable linear map reduces them to a position feature $\psi(x,s_i)\in\R^{m'}$.

The position features are pooled over positions and scales in a frame of the image. Let $(\bar x,\hat\sigma)$ be the intensity frame (\cref{sub-blur}) of the contrast-normalized image. A position $x$ at the scale $s$ has the coordinates
\[
r=\abs{x-\bar x}/\hat\sigma,\qquad \tau=\log(\sqrt s/\hat\sigma),
\]
which are invariant under translations and under a dilation $\Dil_\lambda$ combined with $s\mapsto\lambda^2s$. The pooling uses finitely many bins $b=1,\dots,B$ with centres $(c^r_b,c^\tau_b)$, widths $(\omega_r,\omega_\tau)$ and the memberships
\[
m_b(x,s)=\tfrac12\bigl(1+\cos(\pi d_b/2)\bigr)\ \text{ if } d_b<2,\qquad m_b(x,s)=0\ \text{ otherwise},\qquad d_b^2=\frac{(r-c^r_b)^2}{\omega_r^2}+\frac{(\tau-c^\tau_b)^2}{\omega_\tau^2}.
\]
A foreground gate $g=\sigma\bigl(a_0(\log\tilde\rho-b_0)\bigr)$, with $a_0$ and $b_0$ learnable and $\tilde\rho$ a scale-normalized energy of the jet, and the area $A(s)$ of the sampling cell at the scale $s$ give the weights $w(x,s)=g(x,s)\,A(s)/s$. The pooled vector consists of the weighted means and the relative masses
\[
\Phi_b=\frac{\sum_{x,s}w\,m_b\,\psi}{\sum_{x,s}w\,m_b},\qquad M_b=\frac{\sum_{x,s}w\,m_b}{\sum_{b'}\sum_{x,s}w\,m_{b'}},
\]
and a layer normalization followed by a two-layer perceptron maps it to the output. The grades enter in two places only, the retraction and the foreground gate. A plain average over positions and scales would not be invariant: it is dominated by the finest rungs, which have the most positions, and a shift of the ladder changes which rungs are present. The pooling weights every rung by its area per unit scale and places it by its scale relative to the image.

Three networks are compared (\cref{tab-networks}). The \emph{graded convolutional network} $\GCN$ is the graded network with $\btheta=\bq$. Its \emph{equal-grades twin} $\UCN$ has $\btheta=(1,\dots,1)$, for which $\rhot$ is the Euclidean norm and $\pi_{\btheta}=\pi_{\mathbf1}$ the radial projection, and its \emph{scale-normalized twin} $\SNT$ is $\UCN$ applied to the normalized jet $\tilde f$. The three have the same jet layer on the same ladder and the same layers, pooling and perceptron; they differ in the normalization of the jet and in the foreground energy $\tilde\rho$.

\begin{table}[htbp]
\centering\small
\caption{The three networks. They differ only in the normalization of the jet that the layers read and in the energy $\tilde\rho$ of the foreground gate.}\label{tab-networks}
\begin{tabular}{@{}llll@{}}
\toprule
Network & grades $\btheta$ & input of the layers & foreground energy $\tilde\rho$ \\
\midrule
$\GCN$, graded convolutional network & $\bq=(1,\dots,N)$ & $\pi_{\bq}(f)$ & $\rhoq(\tilde f)=s^{1/2}\rhoq(f)$ \\
$\UCN$, equal-grades twin & $(1,\dots,1)$ & $\pi_{\mathbf1}(f)$ & $s^{1/2}\norm f$ \\
$\SNT$, scale-normalized twin & $(1,\dots,1)$ & $\pi_{\mathbf1}(\tilde f)$ & $\norm{\tilde f}$ \\
\bottomrule
\end{tabular}
\end{table}

\begin{lemma}\label{lem-pool}
The coordinates $r$ and $\tau$ and the memberships $m_b$ are invariant under translations and under the dilation $\Dil_\lambda$ composed with the ladder shift $s\mapsto\lambda^2s$. So are the foreground gates of $\GCN$ and $\SNT$, which are functions of $\tilde f$ (\cref{thm-forced}), and, for the ladder with $\kappa=2$ and the sampling pyramid of \cref{sub-impl}, so is $w$ when $\lambda$ is a power of two, for which the dilation maps the sampling pyramid of a canvas onto that of the dilated canvas, with the rung index shifted. Hence, for this ladder and pyramid, the pooled vectors of $\GCN$ and $\SNT$ are invariant under the dilations by powers of two whenever the sampling grids match and the bins are supported inside the ladder. In the continuous idealization, in which the sums over the pyramid are integrals over all positions and scales, they are invariant under all translations. Under the rotations by multiples of $90^\circ$, which preserve the sampling grid, the bins are invariant, the slots transform by their frequencies, and the pooled vector of the invariant readout is invariant.
\end{lemma}

The full readout, followed by an unconstrained reduction and perceptron, carries no guarantee under rotations; it keeps the orientation. The bins are supported inside the ladder for all images whose radius $\hat\sigma$ keeps $c^\tau_b\pm2\omega_\tau$ within $[\log(\sqrt{s_0}/\hat\sigma),\log(\sqrt{s_{n-1}}/\hat\sigma)]$. Outside this range of sizes a bin is partly empty and the invariance holds only approximately, as for every network on a finite ladder of scales.

By \cref{thm-forced}, $\GCN$ is invariant under the grading action at a fixed scale, $\SNT$ under dilations at corresponding scales, and $\UCN$ under neither, only under the scalar multiplication $f\mapsto cf$, which is contrast. A dilation by $\kappa^{1/2}$ acts by a shift of the ladder index (\cref{lem-bundle}), so a network on $\tilde f$, or on $\pi_{\bq}(f)=\pi_{\bq}(\tilde f)$, with weights shared across the ladder is equivariant under the discrete group generated by $\kappa^{1/2}$, up to the ends of the ladder. The scale-equivariant networks of \cite{WorrallWelling2019, Sosnovik2020} carry such a ladder as well; the continuous group within one rung is described by \cref{lem-octave,prop-octave}.

\begin{proposition}\label{prop-equiv}
Let $\btheta$ be a grading and $\Phi_{\btheta}$ the position features of \cref{sub-networks} built on $\pi_{\btheta}$.
\begin{enumerate}[label=(\roman*)]
\item $\Phi_{\btheta}(\mu\star_{\btheta}f)=\Phi_{\btheta}(f)$ for all $\mu>0$; in particular $\Phi_{\bq}$ is invariant under the dilation action on the jet at a fixed scale, and $\Phi_{(1,\dots,1)}$ is invariant under $f\mapsto cf$, whereas $\pi_{(1,\dots,1)}$ is not constant on the orbits of $\star_{\bq}$.
\item Every slot of frequency $j$ is multiplied by $e^{ij\phi}$ under the rotation by $\phi$, and the invariant readout is rotation-invariant.
\item On the ladder, $\Phi_{\bq}$ evaluated on $\Dil_\lambda I$ at $(x,s)$ equals $\Phi_{\bq}$ evaluated on $I$ at $(x/\lambda,s/\lambda^2)$ for every $\lambda$; for $\lambda=\kappa^{m/2}$ this is a shift of the ladder index by $m$. The same holds for the position features of $\SNT$, which are functions of $\tilde f$. For the ladder with $\kappa=2$ and the sampling pyramid of \cref{sub-impl}, the pooled vectors of $\GCN$ and $\SNT$ are invariant under the dilations by powers of two whenever the sampling grids match and the pooling bins remain inside the ladder, and, in the continuous idealization, under translations (\cref{lem-pool}).
\end{enumerate}
\end{proposition}
\begin{proof}
(i) is \cref{lem-rho} together with the fact that every later layer is a function of $\pi_{\btheta}(f)$; the last assertion holds because the radial projection of $\mu\star_{\bq}f$ differs from that of $f$ unless all nonzero components of $f$ have the same order (\cref{prop-blur}(ii)). (ii) is \eqref{eq-freq} together with the rule for products. (iii) is \cref{lem-bundle} with $\pi_{\bq}(\tilde f)=\pi_{\bq}(f)$, \cref{thm-forced}(i) for $\SNT$, and \cref{lem-pool}.
\end{proof}

In the experiments $N=3$ and the ladder is $s_i=2^{i+1}$, $i=0,\dots,11$, in pixels of the working grid (\cref{sub-impl}). The lift has $m=8$ channels per type and is followed by a gate; one product layer with $w_{\max}=6$, which has $15$ slot types and $120$ slots, is followed by a gate; the full readout is reduced to $m'=16$; the $3\times3$ bins have the centres $c^r\in\{0.5,1,1.5\}$ and $c^\tau\in\{-1,-0.5,0\}$ and the widths $\omega_r=\omega_\tau=0.3$; and the perceptron has $64$ hidden units. Each of the three networks has $21{,}950$ parameters: $7{,}280$ in the equivariant layers, $3{,}856$ in the reduction, $10{,}506$ in the perceptron and $308$ in the layer normalization and the foreground gate.

\subsection{Universality}\label{sub-universal}
\begin{theorem}\label{thm-universal}
Let $F:V^+\smallsetminus0\to\R$ be continuous and invariant under dilations and rotations, and let $K\subset V^+\smallsetminus0$ be compact. For every $\varepsilon>0$ there is a pointwise graded network $\widehat F$ with invariant readout and enough product layers, channels, weight cutoff $w_{\max}$ and perceptron width, such that $\sup_K\abs{F-\widehat F}<\varepsilon$.
\end{theorem}
\begin{proof}
We apply the invariant-polynomial construction of \cite{Yarotsky2022} after the weighted retraction. Write $F=F_0\circ\pi_{\bq}$ with $F_0$ continuous and $\SO(2)$-invariant on the compact weighted sphere $S_{\bq}$; this is possible since $\pi_{\bq}$ is continuous and constant on the dilation orbits. In the harmonic coordinates $c=(c_{k,j})$ the circle acts by \eqref{eq-freq}, a torus action with integer weights, and $\R[V^+]^{\SO(2)}$ is generated by the real and imaginary parts of the finitely many minimal monomials $\prod c_{k,j}^{a_{k,j}}\bar c_{k,j}^{b_{k,j}}$ of total frequency $\sum j(a_{k,j}-b_{k,j})=0$, by Gordan's lemma on the semigroup of exponents \cite{DerksenKemper15}. The invariants of a compact group separate its orbits, which are closed \cite{Hilbert1890, DerksenKemper15}. Let $p=(p_1,\dots,p_r)$ be the generators. The map $p:S_{\bq}/\SO(2)\to\R^r$ is a continuous injection of a compact space into a Hausdorff space, hence a homeomorphism onto its image $Q$, so $F_0=F_1\circ p$ with $F_1$ continuous on $Q$. Extend $F_1$ to $\R^r$ by Tietze's theorem and approximate it uniformly on a ball by a perceptron with one ReLU hidden layer \cite{Cybenko1989, Leshno1993}. It remains to approximate $p$ uniformly on $S_{\bq}$. A monomial $\prod c_{k,j}^{a_{k,j}}\bar c_{k,j}^{b_{k,j}}$ of degree $d$ is a product of $d$ jet slots, of weight $\sum_{k,j}k(a_{k,j}+b_{k,j})$, which lies between $d$ and $Nd$. If $w_{\max}$ is at least the largest weight of the generators, the product layer keeps every partial product, and $d-1$ product layers form the monomial when the gates are replaced by the identity and the layers of kind (a) select channels; the invariant readout returns its real and imaginary parts as slots of frequency zero. A gate $z\mapsto\sigma(\alpha(\abs z-\beta))z$ converges to the identity uniformly on bounded sets as $\beta\to-\infty$, and all slots are bounded on the compact set $S_{\bq}$, so the network with its gates approximates $p$ uniformly on $S_{\bq}$. Since $F_1$ is uniformly continuous on a compact neighbourhood of $Q$, the composite approximates $F_0\circ\pi_{\bq}$ uniformly on $K$.
\end{proof}

\subsection{Pointwise predictors along scale orbits}\label{sub-orbit}
The remark and the theorem that follow concern pointwise predictors on a single jet; the separation for complete classifiers, with foreground gates and pooling, is \cref{thm-local-separation}. Write $x=\mu\star\tilde x$ with $\tilde x=\pi_{\bq}(x)\in S_{\bq}$ and $\mu=\rhoq(x)$ for the orbit coordinates of a jet.

\begin{remark}\label{rem-orbitvar}
Let $x$ be a random jet in $V^+\smallsetminus0$ and $y=F(x)+\epsilon$, where $F$ is dilation-invariant with $\mathbb E F(x)^2<\infty$ and $\epsilon$ is independent of $x$ with mean zero and $\mathbb E\epsilon^2<\infty$, and let $\Risk(\widehat f)=\mathbb E(\widehat f(x)-y)^2$. For every square-integrable predictor $\widehat f$ the orbit average $\bar f(x)=\mathbb E[\widehat f(x)\mid\pi_{\bq}(x)]$ is dilation-invariant, and conditioning on $\pi_{\bq}(x)$ gives
\[
\Risk(\widehat f)-\Risk(\bar f)=\mathbb E\,\Var\bigl(\widehat f(x)\mid\pi_{\bq}(x)\bigr)\ \ge0,
\]
with equality if and only if $\widehat f(x)$ is almost surely a function of $\pi_{\bq}(x)$; no assumption on the law of $\mu$ given $\tilde x$ is needed. Since $F(x)=\mathbb E[y\mid x]$, also $\Risk(\widehat f)-\Risk(F)\ge\mathbb E\,\Var\bigl(\widehat f(x)\mid\pi_{\bq}(x)\bigr)$. A pointwise graded predictor after retraction has orbit variance zero by \cref{prop-equiv}(i). The orbit average $\bar f$ is only measurable, but it is an $L^2$ limit of continuous dilation-invariant functions, since continuous functions are dense in $L^2$ of the law of $\pi_{\bq}(x)$ on the compact space $S_{\bq}$, and those of them that are also rotation-invariant are approximated by graded networks by \cref{thm-universal}.
\end{remark}

\begin{theorem}\label{thm-extrapolation}
Consider pointwise predictors on a single jet, after the retraction and before the foreground gate and the pooling. Fix $\tilde x\in S_{\bq}$ with $\tilde x_N\ne0$ and consider the orbit $\mu\mapsto\mu\star\tilde x$.
\begin{enumerate}[label=(\roman*)]
\item If $\widehat f$ is a network on $V$ with affine layers and activations with finitely many affine pieces, such as a ReLU perceptron or a ReLU convolutional network read through the jet, then $\mu\mapsto\widehat f(\mu\star\tilde x)$ coincides for all sufficiently large $\mu$ with a polynomial of degree at most $N$ in $\mu$; unless that polynomial is constant, $\abs{\widehat f(\mu\star\tilde x)}\to\infty$.
\item If $\widehat f$ is a pointwise $\UCN$, then $\lim_{\mu\to\infty}\widehat f(\mu\star\tilde x)$ exists and depends only on $\tilde x_N/\norm{\tilde x_N}$, with convergence at the rate $\mu^{-1}$, and $\lim_{\mu\to0}\widehat f(\mu\star\tilde x)$ depends only on the lowest nonzero component of $\tilde x$.
\item A pointwise $\GCN$ satisfies $\widehat f(\mu\star\tilde x)=\widehat f(\tilde x)$ for all $\mu>0$.
\end{enumerate}
\end{theorem}
\begin{proof}
(i) A network with affine layers and piecewise-linear activations is affine on each cell of a finite polyhedral partition of $V$. The curve $\mu\mapsto\mu\star\tilde x=(\mu^k\tilde x_k)_k$ is polynomial in $\mu$, so each of the finitely many hyperplanes that bound the cells is crossed at finitely many $\mu$ or contains the curve. Beyond the largest crossing the curve stays in the closure of one cell, on which $\widehat f(\mu\star\tilde x)=c+\sum_k\mu^k\ip{w_k}{\tilde x_k}$. (ii) The radial projection of $\mu\star\tilde x$ converges as $\mu\to\infty$ to $(0,\dots,0,\tilde x_N/\norm{\tilde x_N})$ with error $O(\mu^{-1})$, and as $\mu\to0$ to the unit vector along the lowest nonzero component. The rest of the pointwise network is Lipschitz on the compact sphere, since its linear maps, bounded products, gates and readout are Lipschitz there. (iii) is \cref{prop-equiv}(i).
\end{proof}

At either end of a fibre orbit the pointwise $\UCN$ thus retains a single derivative order. The identity of \cref{rem-orbitvar} uses only that $\pi_{\bq}$ is a complete invariant of the dilation action on the fibre. With the action \eqref{eq-bundleaction} and its complete invariant $\nu$ (\cref{thm-forced}) in place of $\pi_{\bq}$ the same identity holds, and $\GCN$ and $\SNT$ both have orbit variance zero for it. \Cref{thm-extrapolation} concerns the action on the fibre at a fixed scale. There $\SNT$ is the radial projection of a fixed multiple of the jet and behaves like $\UCN$; but a dilation of the image acts by \eqref{eq-bundleaction}, not on the fibre at a fixed scale, and for that action $\SNT$ is invariant. The variance along zooms of images measured in \cref{sec-exp} vanishes for the ideal pooled graded network only under the matching and support conditions of \cref{lem-pool,prop-equiv}, which sampled digits need not meet exactly.

\section{Separation at the zero jet}\label{sec-separation}
By \cref{lem-regularity} the retraction $\pi_{\bq}$ has no continuous extension to the zero jet. In this section we show that at this point the graded quotient separates, with a uniform margin, families of images on which the score gap of every radial probe classifier, and of every classifier continuous in the image, tends to zero. Throughout, $u_1\in V_1$ has norm one and $u_2\in V_2$ is the positive isotropic quadratic form of norm one, whose only nonzero harmonic coordinate is the trace coordinate $h_0$ of \cref{sub-forms}.

\subsection{Probe classifiers and the separation theorem}\label{sub-probe}\label{sub-separation}
Fix finitely many distinct position--scale probes $\mathcal P=\{(x_p,s_p):p=1,\dots,P\}$ with $s_p>0$ and finitely many bins $b=1,\dots,B$, and write $\mathcal J_pI=(f_{1,p},f_{2,p},f_{3,p})$ for the exact Gaussian $3$-jet of positive order of an image $I$ at the probe $p$. Let $\pi$ be one of the normalizations $\pi_{\bq}$, the radial projection $\pi_{\mathbf1}(f)=f/\norm f$ of the raw jet, or the radial projection $\pi_{\mathbf1}\circ\nu$ of the normalized jet, each extended by $\pi(0)=0$.

\begin{definition}\label{def-probe}
A \emph{probe classifier with normalization $\pi$} is a function $H(I)=\Omega\bigl(\Phi(I),M(I)\bigr)$ with
\[
\Phi_b(I)=\frac{\sum_pw_p(I)\,m_{bp}(I)\,\psi\bigl(\pi(\mathcal J_pI)\bigr)}{\sum_pw_p(I)\,m_{bp}(I)+\eta},\qquad
M_b(I)=\frac{\sum_pw_p(I)\,m_{bp}(I)}{\sum_{b',p}w_p(I)\,m_{b'p}(I)},
\]
where $\eta\ge0$, empty means and masses are zero, and
\begin{enumerate}[label=\textup{(P\arabic*)}]
\item $\psi:\pi(V^+)\to\R^d$ is bounded and Lipschitz;
\item the weights $w_p(I)\ge0$ are functions of the block norms $\norm{f_{k,p'}}$, the scales and the intensity frame and mass of $I$, and the memberships $m_{bp}(I)\in[0,1]$ are fixed functions of the probe and the intensity frame;
\item $\Omega$ is Lipschitz on bounded sets.
\end{enumerate}
For given probes, bins and memberships, $\mathcal N_\pi$ denotes the class of probe classifiers with normalization $\pi$.
\end{definition}

Weights that depend on block norms include foreground gates and hard thresholds on them. For images $I_t^\pm$ with labels $\pm1$ let
\[
R_t(H)=\tfrac12\bigl\{(H(I_t^+)-1)^2+(H(I_t^-)+1)^2\bigr\},\qquad
C_t(H)=\tfrac12\bigl\{\log(1+e^{-H(I_t^+)})+\log(1+e^{H(I_t^-)})\bigr\}
\]
be the balanced squared and logistic losses.

\begin{theorem}\label{thm-local-separation}
Let the probes contain $p_0=(0,s_*)$, and let the memberships be those of the pooling of \cref{sub-readout}, with the bins fixed there. There are $t_0>0$ and smooth compactly supported images $I_t^\pm:\R^2\to[0,1]$, $0<t<t_0$, with minimum $0$ and maximum $1$, such that:
\begin{enumerate}[label=\textup{(\roman*)}]
\item the intensity frames and masses of $I_t^\pm$ and their jets at the probes $p\ne p_0$ do not depend on $t$ or on the sign, $\mathcal J_{p_0}I_t^\pm=(tu_1,\pm t^2u_2,0)$, the images converge uniformly to one image $I_0$ as $t\to0$, and $p_0$ has positive membership in some bin;
\item for every $H\in\mathcal N_{\pi_{\mathbf1}}$ there is $K_H<\infty$ with
\[
\abs{H(I_t^+)-H(I_t^-)}\le\frac{2K_Ht}{\sqrt{1+t^2}},\qquad
R_t(H)\ge\Bigl(1-\frac{K_Ht}{\sqrt{1+t^2}}\Bigr)_+^2,\qquad
C_t(H)\ge\log\Bigl(1+e^{-K_Ht/\sqrt{1+t^2}}\Bigr),
\]
and the same holds for every $H\in\mathcal N_{\pi_{\mathbf1}\circ\nu}$ with $\sqrt{1+s_*t^2}$ in place of $\sqrt{1+t^2}$ and $K_H$ multiplied by $\sqrt{s_*}$;
\item there is $H\in\mathcal N_{\pi_{\bq}}$ with $H(I_t^+)=1$ and $H(I_t^-)=-1$ for all $0<t<t_0$.
\end{enumerate}
Consequently
\[
\inf_{H\in\mathcal N_{\pi_{\bq}}}\ \sup_{0<t<t_0}R_t(H)=0,\qquad
\inf_{H\in\mathcal N_{\pi}}\ \sup_{0<t<t_0}R_t(H)=1\qquad(\pi=\pi_{\mathbf1},\ \pi_{\mathbf1}\circ\nu),
\]
the second infimum being attained by $H=0$; for the logistic loss the two infima are $0$ and $\log2$.
\end{theorem}
\begin{proof}
(i) is \cref{lem-local-image-family} below.

(ii) At $p_0$ a classifier with normalization $\pi_{\mathbf1}$ reads
\[
v_t^\pm=\pi_{\mathbf1}(tu_1,\pm t^2u_2,0)=\frac{(u_1,\pm tu_2,0)}{\sqrt{1+t^2}},\qquad \norm{v_t^+-v_t^-}=\frac{2t}{\sqrt{1+t^2}}.
\]
The two jets have the same block norms at every probe and the two images have the same intensity frame and mass, so by (P2) their weights, memberships and masses coincide, and every probe other than $p_0$ contributes the same value of $\psi$. In each weighted mean the coefficient of $p_0$ is at most one, also when $\eta>0$. If $L_\psi$ is a Lipschitz constant of $\psi$, each $\Phi_b$ changes by at most $L_\psi\norm{v_t^+-v_t^-}$, and the pooled vector by at most $\sqrt B\,L_\psi\norm{v_t^+-v_t^-}$. By (P1) the pooled vectors lie in a bounded set, on which $\Omega$ has a Lipschitz constant $L_\Omega$ by (P3); this gives the first bound with $K_H=L_\Omega\sqrt B\,L_\psi$. Write the two scores as $m\pm d$. Then $R_t(H)=m^2+(d-1)^2$ with $\abs d\le K_Ht/\sqrt{1+t^2}$, which gives the second bound, and the convexity of $\ell(z)=\log(1+e^z)$ gives $C_t(H)=\frac12\{\ell(-m-d)+\ell(m-d)\}\ge\ell(-d)$, which gives the third. For $\pi_{\mathbf1}\circ\nu$ the inputs at $p_0$ are $(u_1,\pm\sqrt{s_*}\,tu_2,0)/\sqrt{1+s_*t^2}$, and the argument is the same.

(iii) At $p_0$, $\rhoq(tu_1,\pm t^2u_2,0)^2=\norm{tu_1}^2+\norm{t^2u_2}=2t^2$ and
\[
\pi_{\bq}(tu_1,\pm t^2u_2,0)=\Bigl(\frac{u_1}{\sqrt2},\pm\frac{u_2}{2},0\Bigr),
\]
two points independent of $t$. Take $w_p\equiv1$, $\eta=0$ and $\psi(v)$ the trace coordinate of the second block of $v$, which is bounded and Lipschitz and takes the values $\pm\frac12$ at the two points. The other probes have fixed jets and the memberships are fixed by (i), so the pooled vectors $z_\pm=(\Phi(I_t^\pm),M(I_t^\pm))$ do not depend on $t$, and they differ in a bin in which $p_0$ has positive membership. The affine function $\Omega(z)=2\langle z-(z_++z_-)/2,\,z_+-z_-\rangle/\norm{z_+-z_-}^2$ sends $z_\pm$ to $\pm1$.

By (ii) every classifier in $\mathcal N_{\pi_{\mathbf1}}$ or $\mathcal N_{\pi_{\mathbf1}\circ\nu}$ has $\liminf_{t\to0}R_t\ge1$ and $\liminf_{t\to0}C_t\ge\log2$, and the zero classifier attains these values for every $t$. Replacing $\Omega$ in (iii) by $M\Omega$ gives the logistic loss $\log(1+e^{-M})$, which tends to $0$ as $M\to\infty$.
\end{proof}

The minimization is over score functions, not estimators. Larger weights in a radial classifier change only $K_H$, so the bounds of (ii) hold for every radial classifier, and the separation concerns the supremum over $t$: it gives no positive gap at a fixed $t$, nor in expectation over a fixed law of $t$. It is a separation in squared loss, cross-entropy and margin, not in zero-one error, since a radial classifier with the scores $\pm t$ keeps the correct signs. The family is not a dilation orbit: the parameter $t$ contracts one local jet by its derivative orders while the rest of the measured structure stays fixed.

\begin{remark}\label{rem-networks-probe}
With exact Gaussian jets at finitely many probes, $\GCN$, $\UCN$ and $\SNT$ are probe classifiers with the normalizations $\pi_{\bq}$, $\pi_{\mathbf1}$ and $\pi_{\mathbf1}\circ\nu$. Here $\psi$ is the composition of the equivariant layers and the linear reduction, which is bounded and Lipschitz on the compact image of the normalization; the weights are the foreground gate, a function of block norms and scale, times $A(s)/s$, together with the thresholds of \cref{sub-impl}; the memberships are those of the bins; and $\Omega$ is the layer normalization, with a positive regularizer in its denominator, followed by the perceptron. Moreover $\GCN$ realizes \cref{thm-local-separation}(iii). Set the slope of its foreground gate to zero, so that $g=\frac12$, and retain one real channel of frequency zero equal to the trace coordinate of the second block, which the lift selects and the product layer passes on with zero product coefficients. A gate multiplies a real coordinate by a positive even function of it, so the retained feature is odd and takes opposite nonzero values at the two retracted points. Reduce it to one output coordinate and set another one identically to zero. The two pooled vectors $p_\pm$ share that zero coordinate and their masses; a layer normalization identifies them only if $p_+=ap_-+b\mathbf1$ with $a>0$, the zero coordinate forces $b=0$, and a common positive mass forces $a=1$. Two ReLU units of the perceptron then implement the affine separator, since $r=\ReLU(r)-\ReLU(-r)$. Unused channels and product coefficients are set to zero, so the number of parameters is unchanged.
\end{remark}

\subsection{Construction of the separating images}\label{sub-construction}
Images here are smooth functions on the plane, not pixel arrays, and the probes are exact Gaussian derivatives. Repeated readings are omitted, and the derivatives are written in the basis $\partial^\alpha$, from which the harmonic coordinates differ by an invertible linear change of basis.

\begin{lemma}\label{lem-probe-interpolation}
Let $B\subset\R^2$ be a nonempty open ball. The linear functionals on $C_c^\infty(B)$ consisting of all distinct Gaussian-jet readings
\[
 h\longmapsto\partial^\alpha(g_{s_p}*h)(x_p),\qquad1\le|\alpha|\le3,
\]
together with $\int h$, $\int x_1h$, $\int x_2h$ and $\int|x|^2h$, are linearly independent. Hence their joint map onto its finite-dimensional target is surjective.
\end{lemma}
\begin{proof}
The functionals are integrals against Gaussian derivatives and the four moment polynomials, all analytic on $\R^2$. A linear dependence on $C_c^\infty(B)$ would make a linear combination of these functions vanish on $B$, hence everywhere. Along every ray the Gaussian terms decay faster than any polynomial, so the polynomial part vanishes. Group the remaining terms by centre and variance; each group has the form $P_j(x)\exp(-|x-x_j|^2/(2s_j))$ with $P_j$ a polynomial. Suppose that some group is nonzero, and choose a generic ray $x=rv$ on which every nonzero $P_j$ restricts to a nonzero polynomial and centres with the same variance have distinct projections onto $v$. Among the nonzero groups take the largest variance, and among those the largest projected centre. After division by its Gaussian factor, all other terms decay exponentially relative to its polynomial as $r\to\infty$, a contradiction. Hence every $P_j$ vanishes, and since at a fixed centre and variance distinct Gaussian derivatives have linearly independent Hermite polynomial factors, all coefficients vanish.

For an explicit right inverse, let $\varphi_i$ be the kernels and moment functions, and choose $\chi\in C_c^\infty(B)$ nonnegative and positive on an open set. The matrix $A_{ij}=\int\chi\varphi_i\varphi_j\,dx$ is positive definite, and for prescribed readings $v$ the function $h_v=\chi\sum_j(A^{-1}v)_j\varphi_j$ has exactly those readings.
\end{proof}

\begin{lemma}\label{lem-local-image-family}
Let the probes contain $p_0=(0,s_*)$ and let the memberships be as in \cref{thm-local-separation}. There are $t_0>0$ and $I_0,\zeta_1,\zeta_2\in C_c^\infty(\R^2)$ such that the images $I_t^\pm=I_0+t\zeta_1\pm t^2\zeta_2$, $0<t<t_0$, take values in $[0,1]$ with minimum $0$ and maximum $1$, have the intensity mass, centroid and root-mean-square radius of $I_0$ and its jets at every probe $p\ne p_0$, and satisfy $\mathcal J_{p_0}I_t^\pm=(tu_1,\pm t^2u_2,0)$. Moreover $p_0$ has membership greater than $10^{-3}$ in some bin.
\end{lemma}
\begin{proof}
Take two nonnegative smooth radial bumps, one supported just inside and one just outside the circle $\abs x=\sqrt{2s_*}$. Since
\[
 \Lap g_{s_*}(x)=\Bigl(\frac{|x|^2}{s_*^2}-\frac2{s_*}\Bigr)g_{s_*}(x),
\]
their Gaussian-smoothed Laplacians at the origin have opposite signs, and a positive linear combination of the two bumps has zero smoothed Laplacian there. Normalizing its maximum to one gives $I_0$. Radial symmetry kills the orders one and three and makes the Hessian isotropic, so the whole $3$-jet of positive order of $I_0$ at $(0,s_*)$ vanishes. For bumps close enough to the circle, the root-mean-square radius is close to $\sqrt{2s_*}$, so that $p_0$ has $r=0$ and $\tau$ close to $-\frac12\log2$. In the bin with centre $(\frac12,-\frac12)$ and widths $0.3$ its membership then approaches
\[
\frac12\left(1+\cos\left[\frac\pi2
\sqrt{\left(\frac{1/2}{0.3}\right)^2+
\left(\frac{1/2-(\log2)/2}{0.3}\right)^2}\right]\right)>0.03.
\]

Choose an open ball $B$ on which $\delta<I_0<1-\delta$ for some $\delta>0$. By \cref{lem-probe-interpolation} there are $\zeta_1,\zeta_2\in C_c^\infty(B)$ whose four moments vanish, whose jets vanish at every probe $p\ne p_0$, and whose jets at $p_0$ are $\mathcal{J}_{p_0}\zeta_1=(u_1,0,0)$ and $\mathcal{J}_{p_0}\zeta_2=(0,u_2,0)$. For $t_0$ small enough the images $I_t^\pm=I_0+t\zeta_1\pm t^2\zeta_2$ take values in $[0,1]$; outside $B$ they coincide with $I_0$, so they keep its minimum $0$ and maximum $1$ and the contrast normalization leaves them unchanged. The vanishing moments fix the intensity mass, centroid and root-mean-square radius, and the linearity of Gaussian differentiation gives the prescribed jets.
\end{proof}

\subsection{Continuous classifiers and regularization}\label{sub-continuous}
\begin{corollary}\label{cor-continuous-cnn-separation}
Let $\mathcal D\subset\R^2$ be a compact canvas containing the supports of the images $I_t^\pm=I_0+t\zeta_1\pm t^2\zeta_2$ of \cref{lem-local-image-family}, and regard an image as an element of $C(\mathcal D)$ with the uniform norm. If a function $H:C(\mathcal D)\to\R$ is continuous at $I_0$, then
\[
\liminf_{t\downarrow0}R_t(H)\ge1,\qquad\liminf_{t\downarrow0}C_t(H)\ge\log2.
\]
If $H$ is $L_H$-Lipschitz on the family, then
\[
R_t(H)\ge(1-L_H\norm{\zeta_2}_\infty t^2)_+^2,\qquad C_t(H)\ge\log\bigl(1+\exp[-L_H\norm{\zeta_2}_\infty t^2]\bigr).
\]
This holds in particular for every convolutional network built from continuous convolutions, continuous activations, finite max or average pooling and normalizations with a positive regularizer in the denominator.
\end{corollary}
\begin{proof}
$I_t^\pm\to I_0$ uniformly, so continuity gives $H(I_t^+)-H(I_t^-)\to0$, and the two-score inequalities in the proof of \cref{thm-local-separation} give the limits. In the Lipschitz case $\abs{H(I_t^+)-H(I_t^-)}\le L_H\norm{I_t^+-I_t^-}_\infty=2L_H\norm{\zeta_2}_\infty t^2$.
\end{proof}

The labels have different limits along two paths to the same image $I_0$, so the ideal graded classifier is discontinuous at $I_0$. The corollary applies equally to a classifier that reads the exact Gaussian probes, which are continuous functionals of the image, and to a graded classifier with a regularized norm and without hard thresholds. Hard thresholds need not be continuous; the next proposition covers them.

\begin{proposition}\label{prop-regularized-collapse}
Fix $\varepsilon>0$ and let $\pi_{\bq,\varepsilon}$ be the retraction with the blockwise regularized norm
\[
\rho_{\bq,\varepsilon}(f)^2=\sum_{k=1}^3\bigl(\norm{f_k}^2+\varepsilon^2\bigr)^{1/k},\qquad
(\pi_{\bq,\varepsilon}f)_k=\frac{f_k}{\rho_{\bq,\varepsilon}(f)^k},
\]
which is the norm of the implementation (\cref{sub-impl}). On the images of \cref{thm-local-separation} put $r_\varepsilon(t)^2=(t^2+\varepsilon^2)+(t^4+\varepsilon^2)^{1/2}+\varepsilon^{2/3}$. Every $H\in\mathcal N_{\pi_{\bq,\varepsilon}}$ satisfies
\[
\abs{H(I_t^+)-H(I_t^-)}\le\frac{2K_Ht^2}{r_\varepsilon(t)^2}\le2K_H\varepsilon^{-2/3}t^2,
\]
hence $R_t(H)\ge(1-a_t)_+^2$ and $C_t(H)\ge\log(1+e^{-a_t})$ with $a_t=K_Ht^2/r_\varepsilon(t)^2\to0$. The minimax squared and logistic losses of $\mathcal N_{\pi_{\bq,\varepsilon}}$ on the family are $1$ and $\log2$.
\end{proposition}
\begin{proof}
The two jets at $p_0$ have the common regularized norm $r_\varepsilon(t)$, so
\[
\norm{\pi_{\bq,\varepsilon}(tu_1,t^2u_2,0)-\pi_{\bq,\varepsilon}(tu_1,-t^2u_2,0)}=\frac{2t^2}{r_\varepsilon(t)^2}\le2\varepsilon^{-2/3}t^2.
\]
Every block of a regularized retraction has norm at most one, so (P1) applies on a bounded set, and the rest is the proof of \cref{thm-local-separation}(ii): the weights, memberships and masses of the two images coincide, even where a threshold makes them jump as functions of $t$.
\end{proof}

The unregularized norm of the selected jet is $\rhoq=\sqrt2\,t$, the regularized one is at least $\varepsilon^{1/3}$, and the two are comparable for $t$ of order $\varepsilon^{1/3}$; we prove no advantage on a range of $t$ bounded away from zero. What \cref{thm-local-separation} isolates is the contribution of the fibre quotient: along the contraction the two classes converge to one image, the score gap of every classifier continuous in the image and of every radial probe classifier tends to zero, and the graded retraction sees two fixed jets for every $t$. The robustness measured in \cref{sec-exp} arises away from this singular point.

\section{Classification under zoom and unknown blur}\label{sec-exp}
We test \cref{thm-forced} and \cref{prop-blur} on the classification of handwritten digits and, with the same networks, of images of garments. We compare three networks with the same architecture, the same number of parameters and the same training protocol; \cref{tab-networks} gives their jet normalizations and foreground energies, and the foreground gates are then exchanged to measure the effect of each choice.

\subsection{Protocol}\label{sub-models}
The digits are MNIST images on a canvas of $29$ pixels, or $57$ pixels when they are enlarged or blurred, contrast normalized as in \cref{sub-group}. A network is trained on $10{,}000$ unscaled training digits and validated on $2{,}000$ others, the first and the last digits of a random permutation of the MNIST training set; three permutations give three training splits. It is tested on the first $5{,}000$ MNIST test digits, presented at the zooms $0.5$, $0.7$, $1$, $1.4$ and $2$ \cite{WorrallWelling2019}; blurred by a sampled Gaussian of standard deviation $\sigma_b\in\{0.5,1,1.5,2\}$ input pixels and contrast normalized again; at half resolution, by an antialiased reduction and a bicubic enlargement; and at the zooms $0.5$ and $2$ with $\sigma_b=1$. For the variance of the output along zoom orbits, $200$ test digits are presented at eleven zooms, log-uniform in $[1/2,2]$ and in $[1/\sqrt2,\sqrt2]$. No network sees a blurred digit in training.

We compare $\GCN$ with its twins $\UCN$ and $\SNT$ (\cref{tab-networks}), each with $21{,}950$ parameters, and with an ordinary convolutional network with three convolutional layers, global average pooling and $22{,}826$ parameters. The latter is applied to inputs rescaled about the canvas centre so that their intensity radius $\hat\sigma$ equals the median radius of the training digits, which gives it the global size invariance that the pooling frame gives the graded networks; we call it the size-normalized network. All networks are trained with Adam and a one-cycle schedule, with peak learning rate $2\cdot10^{-3}$, batch $64$ and weight decay $10^{-4}$, for $12$ epochs ($1{,}884$ steps), and the checkpoint with the best validation accuracy is kept. $\GCN$ and its twins are trained on the three splits; the networks with exchanged gates and the size-normalized network are trained on split $1$.

Two networks are compared on the same test digits by the exact two-sided McNemar test on the per-digit predictions \cite{McNemar1947}, with the Holm adjustment \cite{Holm1979} over the ten conditions of \cref{tab-blur-v7} within each split. The binomial standard error of an accuracy near $95\%$ on $5{,}000$ digits is $0.3$ points.

The gradings and ablations of \cref{tab-grades,tab-abl} come from earlier runs on MNIST-Scale, digits rescaled by a factor uniform in $[0.3,1]$ \cite{SohnLee2012, Sosnovik2020} and tested on the official MNIST test digits, so that they do not reproduce the protocol of that benchmark. In these runs the validation digits followed the training digits in the permutation, the rescalings of the test digits depend on the training size, the per-digit predictions were not recorded, and each configuration was trained once; we interpret only differences of several points.

\subsection{Scale}\label{sub-exp-zoom}
\begin{figure}[htbp]
\centering
\includegraphics[width=\textwidth]{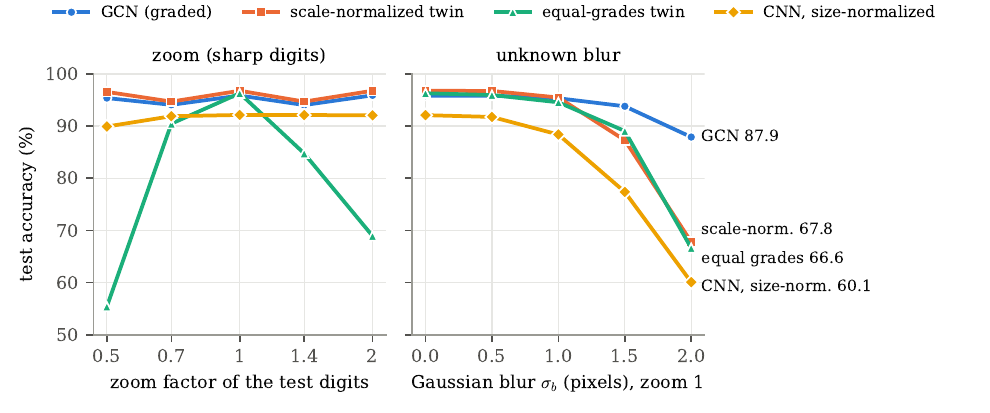}
\caption{The graded network ($\GCN$) against three baselines, all trained on $10{,}000$ sharp unscaled MNIST digits and tested on $5{,}000$ digits. Left: test digits zoomed by the factor shown. Right: test digits at their original size, blurred by a Gaussian of standard deviation $\sigma_b$ pixels. The scale-normalized twin is the same network on scale-normalized derivatives, the equal-grades twin the same network with all grades equal, and the last baseline an ordinary convolutional network on inputs rescaled to a fixed size. Split $1$; means over three splits in \cref{tab-graded-v7,tab-blur-v7}.}\label{fig-main}
\end{figure}

\Cref{fig-main} (left) and \cref{tab-graded-v7} test \cref{thm-forced} under zoom. The zooms $0.5$ and $2$ shift the ladder by two rungs, under which $\GCN$ and $\SNT$ are invariant (\cref{prop-equiv}); the zooms $0.7$ and $1.4$ are not shifts of the ladder. Trained at the single scale $1$, $\GCN$ keeps $95.4\%$ and $96.0\%$ at the zooms $0.5$ and $2$ on average, against $96.0\%$ at the training scale, and loses about $2.7$ points at $0.7$ and $1.4$; its lowest accuracy over the five zooms and three splits is $92.3\%$. The same network with equal grades loses $34$ and $17$ points at $0.5$ and $2$ on average, and up to $41$ in one split; on split $1$, at the zoom $0.5$, $\GCN$ is right and $\UCN$ wrong on $2{,}046$ digits, and the reverse holds on $47$. Along the zoom orbits the mean variance of the softmax output is $0.017$--$0.024$ for $\GCN$ and $0.093$--$0.144$ for $\UCN$, and the predicted label is constant along the orbit for $85$--$89\%$ and $33$--$52\%$ of the digits. This is the behaviour that \cref{rem-orbitvar,thm-extrapolation} describe pointwise. Exchanging the foreground gates shows that the retraction, not the gate, carries the invariance: $\GCN$ with the gate of $\UCN$ is as flat as with its own, while $\UCN$ with the graded gate still loses $20$ points at both shifts of the ladder.

\begin{table}[htbp]
\centering\footnotesize
\caption{Zoom. Networks trained on $10{,}000$ sharp unscaled digits, tested on $5{,}000$ test digits resampled by the zoom factor: accuracy (\%), mean variance of the softmax output over eleven zooms in $[1/2,2]$ and in $[1/\sqrt2,\sqrt2]$ ($200$ digits), and percentage of digits whose predicted label is constant over $[1/2,2]$. Upper block: mean $\pm$ half-range over three training splits; lower block: split $1$. $\pi_{\mathbf 1}$ is the radial projection, $\tilde f=s^{1/2}\star f$ the normalized jet.}\label{tab-graded-v7}
\setlength{\tabcolsep}{1.8pt}
\begin{tabular}{@{}lllrrrrrrrr@{}}
\toprule
Model & retr. & gate & $0.5$ & $0.7$ & $1$ & $1.4$ & $2$ & var.\ $[\frac12,2]$ & var.\ $[\frac1{\sqrt2},\sqrt2]$ & const. \\
\midrule
$\GCN$ & $\pi_{\bq}(f)$ & $\rhoq(\tilde f)$ & $95.4{\scriptstyle\pm0.0}$ & $93.4{\scriptstyle\pm0.8}$ & $96.0{\scriptstyle\pm0.1}$ & $93.2{\scriptstyle\pm0.9}$ & $96.0{\scriptstyle\pm0.1}$ & $0.020{\scriptstyle\pm0.003}$ & $0.022{\scriptstyle\pm0.003}$ & $87{\scriptstyle\pm2}$ \\
$\SNT$ & $\pi_{\mathbf 1}(\tilde f)$ & $\norm{\tilde f}$ & $96.2{\scriptstyle\pm0.4}$ & $94.5{\scriptstyle\pm0.3}$ & $96.8{\scriptstyle\pm0.2}$ & $94.5{\scriptstyle\pm0.4}$ & $96.8{\scriptstyle\pm0.2}$ & $0.019{\scriptstyle\pm0.002}$ & $0.018{\scriptstyle\pm0.003}$ & $89{\scriptstyle\pm1}$ \\
$\UCN$ & $\pi_{\mathbf 1}(f)$ & $s^{1/2}\norm f$ & $62.6{\scriptstyle\pm9.5}$ & $91.1{\scriptstyle\pm0.7}$ & $96.2{\scriptstyle\pm0.1}$ & $87.9{\scriptstyle\pm2.5}$ & $79.6{\scriptstyle\pm8.5}$ & $0.114{\scriptstyle\pm0.025}$ & $0.045{\scriptstyle\pm0.009}$ & $43{\scriptstyle\pm10}$ \\
\midrule
$\GCN$, equal-grades gate & $\pi_{\bq}(f)$ & $s^{1/2}\norm f$ & 95.5 & 94.3 & 95.9 & 94.1 & 95.9 & 0.017 & 0.018 & 89 \\
$\UCN$, graded gate & $\pi_{\mathbf 1}(f)$ & $\rhoq(\tilde f)$ & 76.1 & 91.6 & 96.5 & 89.6 & 76.3 & 0.090 & 0.042 & 52 \\
CNN, size-normalized & -- & -- & 89.9 & 91.9 & 92.1 & 92.1 & 92.1 & 0.004 & 0.002 & 93 \\
\bottomrule
\end{tabular}
\end{table}

The scale-normalized twin is invariant as well, as \cref{thm-forced}(i) predicts for the bundle quotient, and it is $0.8$ to $1.3$ points more accurate than $\GCN$ on average. The size-normalized network is flat across zooms, at $89.9$--$92.1\%$, about four points below $\GCN$. Trained on MNIST-Scale, $\GCN$ reaches $95.6$--$95.7\%$ in distribution and stays within $0.4$ points of it at the zooms $0.5$ and $2$, while $\UCN$, at $95.4$--$95.7\%$ in distribution, falls to $81.9$--$85.5\%$ at the zoom $2$, which lies outside its training scales; $\SNT$ reaches $95.8$--$96.5\%$ and is flat, and the size-normalized network reaches $86.96\%$ in distribution and $88.1$--$88.4\%$ at the five zooms (split $1$). In the earlier runs, an ordinary convolutional network trained on unscaled digits reaches $94.9\%$ at its training scale and $36.9\%$ and $33.1\%$ at the zooms $0.5$ and $2$; trained with scale augmentation by a factor log-uniform in $[\frac12,2]$ for five times the epochs, it reaches $92.8\%$, $75.4\%$ and $86.1\%$, while $\GCN$ reaches $95.7\%$, $95.5\%$ and $95.7\%$ without augmentation.

\begin{figure}[htbp]
\centering
\includegraphics[width=0.6\textwidth]{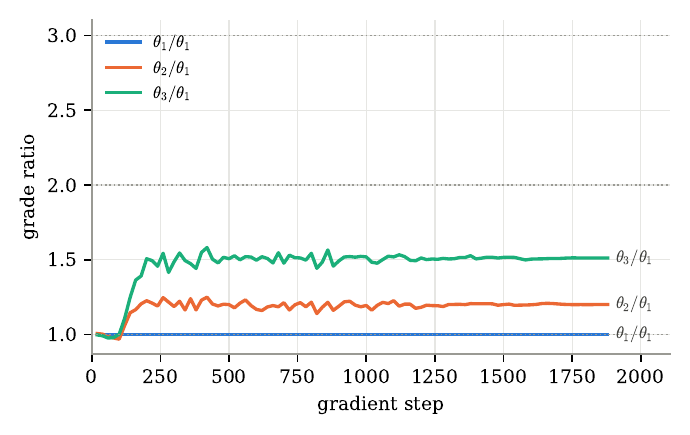}
\caption{The ratios $\theta_k/\theta_1$ of the learnable grades during training, started at the equal grading $(1,1,1)$ with the grade learning rate at five times the base rate; the dotted lines are the ratios $1$, $2$, $3$ of the derivative orders.}\label{fig-theta}
\end{figure}

\Cref{tab-grades} tests parts (iv) and (v) of \cref{thm-forced}. The gradings $(3,1,2)$ and $(1,3,2)$ are exactly invariant under their own action but not under the action of dilations. At the zoom $2$, outside the training scales of MNIST-Scale, they keep $83.1\%$ and $92.3\%$, against $95.5\%$ for the correct grading and $82.2\%$ for equal grades. A learnable grading $\btheta=e^{\boldsymbol\varphi}$ started at $(1,1,1)$, with the grade parameters at five and at twenty times the base learning rate, keeps $89.5\%$ and $89.4\%$. Its grades move toward the ratios $1\!:\!2\!:\!3$ of the derivative orders, to $1\!:\!1.20\!:\!1.51$ and $1\!:\!1.22\!:\!1.49$ at the best-validation checkpoint, and stop short of them (\cref{fig-theta}).

\begin{table}[htbp]
\centering\small
\caption{Gradings. Test accuracy (\%) in distribution and at the zoom factors $0.5$ and $2.0$ (test digits resampled), and orbit variance in $[1/2,2]$, for the correct, equal, wrong and learned gradings, all trained on MNIST-Scale. The learned gradings are those of the best-validation checkpoint, reported up to the common scale that leaves $\theta_1=1$.}\label{tab-grades}
\begin{tabular}{@{}llrrrr@{}}
\toprule
Model & $\btheta$ & test & zoom $0.5$ & zoom $2.0$ & orbit var. \\
\midrule
$\GCN$ & $(1,2,3)$ & 95.3 & 95.7 & 95.5 & 0.0110 \\
$\UCN$ & $(1,1,1)$ & 95.3 & 95.9 & 82.2 & 0.0338 \\
wrong & $(3,1,2)$ & 93.8 & 94.1 & 83.1 & 0.0356 \\
wrong & $(1,3,2)$ & 95.0 & 95.3 & 92.3 & 0.0246 \\
learned & $(1.00, 1.20, 1.51)$ & 95.6 & 96.2 & 89.5 & 0.0217 \\
learned, grade l.r.\ $\times20$ & $(1.00, 1.22, 1.49)$ & 95.7 & 96.3 & 89.4 & 0.0205 \\
\bottomrule
\end{tabular}
\end{table}

\Cref{tab-abl} reports ablations from the same earlier runs. The network on the $2$-jet, with $15{,}702$ parameters, reaches $94.3\%$ in distribution, one point below the network on the $3$-jet, and the network without the product layer, with $12{,}230$ parameters, reaches $92.1\%$; both stay flat across zoom. With the invariant readout $\GCN$ is invariant under rotations by $90^\circ$, and nearly so at $30^\circ$ and $60^\circ$, where interpolation enters, at a cost of $6.6$ points in distribution; for upright digits the orientation carries class information.

\begin{table}[htbp]
\centering\small
\caption{Ablations and rotation, trained on MNIST-Scale: test accuracy (\%) in distribution, at zoom $2.0$, and on the test digits rotated by $30^\circ$, $60^\circ$, $90^\circ$.}\label{tab-abl}
\begin{tabular}{@{}lrrrrr@{}}
\toprule
Model & test & zoom $2.0$ & $30^\circ$ & $60^\circ$ & $90^\circ$ \\
\midrule
$\GCN$, $N=3$, full readout & 95.3 & 95.5 & 71.7 & 24.8 & 12.2 \\
$\GCN$, $N=2$ & 94.3 & 94.3 & 67.3 & 18.3 & 9.3 \\
$\GCN$, no product layer & 92.1 & 92.5 & 67.7 & 22.6 & 11.7 \\
$\GCN$, invariant readout & 88.7 & 87.1 & 86.1 & 86.1 & 87.1 \\
$\UCN$, full readout & 95.3 & 82.2 & 66.6 & 15.7 & 8.5 \\
$\UCN$, invariant readout & 88.1 & 51.1 & 81.6 & 81.7 & 81.9 \\
CNN & 89.1 & 32.0 & 40.8 & 8.8 & 7.4 \\
\bottomrule
\end{tabular}
\end{table}

\subsection{Unknown blur}\label{sub-exp-blur}
\Cref{fig-main} (right) and \cref{tab-blur-v7} test \cref{prop-blur} on the networks of \cref{tab-graded-v7}, which were trained on sharp digits only. Averaged over the three splits, $\GCN$ loses $0.4$ points at $\sigma_b=1$, $2.1$ at $\sigma_b=1.5$ and $7.6$ at $\sigma_b=2$. The scale-normalized twin, identical except for the normalization and the foreground energy, loses $1.7$, $9.8$ and $30.6$ points, and the equal-grades twin and the size-normalized network fall with it. At $\sigma_b=2$, $\GCN$ is right and $\SNT$ wrong on $1{,}155$, $1{,}394$ and $1{,}166$ digits in the three splits, and the reverse holds on $150$, $107$ and $132$. Half-size digits blurred by one pixel, which model a smaller object seen through the same optics, give the same picture: $86.3\%$ for $\GCN$ against $62.6\%$ for $\SNT$, $73.5\%$ for $\UCN$ and $55.1\%$ for the size-normalized network. After the Holm adjustment within each split, $\GCN$ is the more accurate at $\sigma_b=1.5$, at $\sigma_b=2$ and on the blurred half-size digits in all three splits ($p<10^{-51}$), and at $\sigma_b=1$ in two of them; $\SNT$ is the more accurate on sharp digits in all three, and under some of the mild conditions. At the zoom $2$ a blur of one pixel is small compared with the enlarged strokes and changes nothing.

\begin{table}[htbp]
\centering\scriptsize
\caption{Unknown blur. The networks of \cref{tab-graded-v7}, trained on sharp digits and not retrained, tested on the same $5{,}000$ test digits on the canvas $57$: blurred by a Gaussian of standard deviation $\sigma_b$ input pixels and contrast normalized again; with the resolution halved (antialiased reduction by two, bicubic enlargement); and at the zooms $0.5$ and $2$, sharp and with $\sigma_b=1$. Accuracy in \%, mean $\pm$ half-range over three training splits (the CNN: split $1$). Last row: the number of splits in which $\GCN$ or $\SNT$ is the more accurate by the exact two-sided McNemar test on the same digits, Holm-adjusted over the ten conditions \cite{Holm1979}, at $p<0.05$.}\label{tab-blur-v7}
\setlength{\tabcolsep}{2pt}
\begin{tabular}{@{}lrrrrrrrrrr@{}}
\toprule
& \multicolumn{5}{c}{zoom $1$, blur $\sigma_b$} & res. & \multicolumn{2}{c}{zoom $0.5$, $\sigma_b$} & \multicolumn{2}{c}{zoom $2$, $\sigma_b$} \\
\cmidrule(lr){2-6}\cmidrule(lr){8-9}\cmidrule(l){10-11}
Model & $0$ & $0.5$ & $1$ & $1.5$ & $2$ & $\times\frac12$ & $0$ & $1$ & $0$ & $1$ \\
\midrule
$\GCN$ & $96.0{\scriptstyle\pm0.2}$ & $95.9{\scriptstyle\pm0.1}$ & $95.6{\scriptstyle\pm0.2}$ & $93.9{\scriptstyle\pm0.3}$ & $88.4{\scriptstyle\pm0.5}$ & $95.4{\scriptstyle\pm0.1}$ & $95.4{\scriptstyle\pm0.0}$ & $86.3{\scriptstyle\pm1.4}$ & $96.0{\scriptstyle\pm0.1}$ & $95.9{\scriptstyle\pm0.2}$ \\
$\SNT$ & $96.8{\scriptstyle\pm0.2}$ & $96.7{\scriptstyle\pm0.2}$ & $95.1{\scriptstyle\pm0.4}$ & $87.0{\scriptstyle\pm0.3}$ & $66.2{\scriptstyle\pm2.7}$ & $96.0{\scriptstyle\pm0.4}$ & $96.2{\scriptstyle\pm0.4}$ & $62.6{\scriptstyle\pm3.1}$ & $96.8{\scriptstyle\pm0.2}$ & $96.6{\scriptstyle\pm0.3}$ \\
$\UCN$ & $96.2{\scriptstyle\pm0.1}$ & $96.0{\scriptstyle\pm0.1}$ & $94.9{\scriptstyle\pm0.4}$ & $90.7{\scriptstyle\pm1.4}$ & $76.0{\scriptstyle\pm7.1}$ & $95.1{\scriptstyle\pm0.1}$ & $62.6{\scriptstyle\pm9.5}$ & $73.5{\scriptstyle\pm3.6}$ & $79.6{\scriptstyle\pm8.5}$ & $77.9{\scriptstyle\pm9.6}$ \\
CNN, size-norm. & 92.1 & 91.8 & 88.4 & 77.4 & 60.1 & 89.9 & 89.9 & 55.1 & 92.1 & 91.7 \\
\midrule
more accurate & $\SNT$ 3/3 & $\SNT$ 2/3 & $\GCN$ 2/3 & $\GCN$ 3/3 & $\GCN$ 3/3 & $\SNT$ 1/3 & $\SNT$ 2/3 & $\GCN$ 3/3 & $\SNT$ 3/3 & $\SNT$ 2/3 \\
\bottomrule
\end{tabular}
\end{table}

\Cref{tab-blur-mech-v7} separates the causes on split $1$. \Cref{prop-blur}(ii) holds exactly only without the second contrast normalization, which multiplies a blurred digit by $c_b=1/\sup(g_b*I)$; over the first $2{,}000$ test digits the median of $c_b$ is $1.13$, $1.37$ and $1.66$ at $\sigma_b=1$, $1.5$ and $2$, and its maximum is $4.0$. Without this normalization $\GCN$ keeps $88.5\%$ at $\sigma_b=2$ and $84.9\%$ on the blurred half-size digits, slightly more than with it, while $\SNT$ keeps $64.7\%$ and $59.5\%$. The contrast factor that the retraction keeps (\cref{prop-blur}(iii)) therefore does not limit $\GCN$, and the one that the radial projection removes does not explain the failure of $\SNT$. With the foreground gates exchanged, $\GCN$ with the gate of $\UCN$ keeps $88.0\%$ and $84.8\%$, as with its own, while $\UCN$ with the graded gate of $\GCN$ falls to $64.5\%$ and $72.7\%$, as with its own. The gain of $\GCN$ over $\UCN$ thus persists when the foreground gates are exchanged: it is carried by the retraction and not by the gate. It does not come from the grading factor of \cref{prop-blur}(ii), which the raw jet read by $\UCN$ does not acquire (\cref{tab-local}); what $\UCN$ lacks is the invariance under dilations, and we have not tested whether this accounts for its loss. The gain of $\GCN$ over $\SNT$ is consistent with the order-dependent distortion of the scale-normalized jet in \cref{prop-blur}(ii): with $\mu^2=s/(s+b)$, the factor $\mu\star$ is far from one once the blur variance $b=4\sigma_b^2$ working pixels$^2$ is comparable to the finest rung $s_0=2$, and on real digits a blur of $0.5$ or $1$ input pixel already changes the normalized jets by up to $24\%$, against at most $1.3\%$ for the retracted jets (\cref{sub-verification}).

\begin{table}[htbp]
\centering\footnotesize
\caption{The mechanism of the blur robustness (split $1$): accuracy (\%) on the $5{,}000$ test digits of \cref{tab-blur-v7}, blurred by $\sigma_b$ input pixels at zoom $1$, and at the zooms $\frac12$ and $2$ with $\sigma_b=1$. Without the second contrast normalization the jet of the blurred image is exactly the jet of the sharp image at the scale $s+b$ (\cref{prop-blur}); exchanging the foreground gates separates the retraction from the gate.}\label{tab-blur-mech-v7}
\setlength{\tabcolsep}{4pt}
\begin{tabular}{@{}lrrrrrrr@{}}
\toprule
& \multicolumn{5}{c}{zoom $1$, blur $\sigma_b$} & \multicolumn{2}{c}{zoom, $\sigma_b$} \\
\cmidrule(lr){2-6}\cmidrule(l){7-8}
Model & $0$ & $0.5$ & $1$ & $1.5$ & $2$ & $\frac12$, $1$ & $2$, $1$ \\
\midrule
\multicolumn{8}{@{}l}{\emph{without the second contrast normalization}} \\
$\GCN$ & 95.8 & 95.8 & 95.4 & 93.6 & 88.5 & 84.9 & 95.8 \\
$\SNT$ & 96.8 & 96.7 & 95.4 & 84.4 & 64.7 & 59.5 & 96.7 \\
\midrule
\multicolumn{8}{@{}l}{\emph{gates exchanged, contrast normalized as in \cref{tab-blur-v7}}} \\
$\GCN$, equal-grades gate & 95.9 & 96.0 & 95.5 & 93.6 & 88.0 & 84.8 & 95.9 \\
$\UCN$, graded gate & 96.5 & 96.4 & 95.2 & 87.0 & 64.5 & 72.7 & 74.6 \\
\bottomrule
\end{tabular}
\end{table}

\subsection{Fashion-MNIST}\label{sub-exp-fashion}
The comparisons of \cref{sub-exp-zoom,sub-exp-blur} are repeated on Fashion-MNIST \cite{XiaoRasulVollgraf2017}, images of garments, shoes and bags in ten classes, in the format of MNIST. Nothing else is changed: the networks, the canvases, the three permutations that define the training splits, the schedule, the test conditions and the tests are those of \cref{sub-models}, and no parameter was adjusted. These images differ from digits in two respects. They have texture and interior grey levels, where a digit is a stroke, and they fill their frame: on average $4.8\%$ of the intensity of a test image lies within two pixels of the border of the canvas $29$, against $0.03\%$ for a digit.

\begin{table}[htbp]
\centering\scriptsize
\caption{Fashion-MNIST. The networks and the protocol of \cref{tab-graded-v7,tab-blur-v7}, trained on $10{,}000$ sharp unscaled images and tested on the first $5{,}000$ test images. Left block: test images resampled by the zoom factor, on the canvas $29$ up to the zoom $1$ and on the canvas $57$ above it. Right block: on the canvas $57$, at the zoom $1$ blurred by a Gaussian of standard deviation $\sigma_b$ input pixels and contrast normalized again, and at the zoom $\frac12$ with $\sigma_b=1$. Accuracy in \%, mean $\pm$ half-range over three training splits (the CNN: split $1$). Last rows: the number of splits in which one network of the pair is the more accurate by the exact two-sided McNemar test on the same images, Holm-adjusted over the ten columns \cite{Holm1979}, at $p<0.05$; ``$1/3$ each'' means that each network is the more accurate in one split.}\label{tab-fashion}
\setlength{\tabcolsep}{2pt}
\begin{tabular}{@{}lrrrrrrrrrr@{}}
\toprule
& \multicolumn{5}{c}{zoom, sharp} & \multicolumn{4}{c}{zoom $1$, blur $\sigma_b$} & zoom $\frac12$ \\
\cmidrule(lr){2-6}\cmidrule(lr){7-10}\cmidrule(l){11-11}
Model & $0.5$ & $0.7$ & $1$ & $1.4$ & $2$ & $0$ & $1$ & $1.5$ & $2$ & $\sigma_b=1$ \\
\midrule
$\GCN$ & $75.6{\scriptstyle\pm2.1}$ & $65.9{\scriptstyle\pm0.9}$ & $83.7{\scriptstyle\pm0.5}$ & $65.6{\scriptstyle\pm1.2}$ & $83.7{\scriptstyle\pm0.4}$ & $77.8{\scriptstyle\pm1.0}$ & $74.7{\scriptstyle\pm2.1}$ & $70.8{\scriptstyle\pm3.2}$ & $63.0{\scriptstyle\pm4.2}$ & $61.0{\scriptstyle\pm4.2}$ \\
$\SNT$ & $75.5{\scriptstyle\pm1.8}$ & $63.4{\scriptstyle\pm1.7}$ & $83.9{\scriptstyle\pm0.6}$ & $64.0{\scriptstyle\pm1.7}$ & $83.7{\scriptstyle\pm0.5}$ & $77.7{\scriptstyle\pm1.4}$ & $74.3{\scriptstyle\pm1.2}$ & $67.7{\scriptstyle\pm1.3}$ & $54.1{\scriptstyle\pm2.7}$ & $52.1{\scriptstyle\pm2.3}$ \\
$\UCN$ & $41.9{\scriptstyle\pm1.4}$ & $55.6{\scriptstyle\pm2.5}$ & $81.4{\scriptstyle\pm0.6}$ & $45.1{\scriptstyle\pm3.6}$ & $52.6{\scriptstyle\pm2.7}$ & $74.6{\scriptstyle\pm0.8}$ & $73.4{\scriptstyle\pm1.0}$ & $68.9{\scriptstyle\pm1.9}$ & $59.3{\scriptstyle\pm2.9}$ & $43.4{\scriptstyle\pm5.5}$ \\
CNN, size-norm. & 70.0 & 73.7 & 74.8 & 74.0 & 74.3 & 74.9 & 67.4 & 57.5 & 51.2 & 50.9 \\
\midrule
$\GCN$ vs.\ $\SNT$ & 1/3 each & $\GCN$ 2/3 & -- & $\GCN$ 2/3 & -- & -- & 1/3 each & $\GCN$ 1/3 & $\GCN$ 3/3 & $\GCN$ 2/3 \\
$\GCN$ vs.\ $\UCN$ & $\GCN$ 3/3 & $\GCN$ 3/3 & $\GCN$ 3/3 & $\GCN$ 3/3 & $\GCN$ 3/3 & $\GCN$ 3/3 & $\GCN$ 1/3 & $\GCN$ 1/3 & $\GCN$ 2/3 & $\GCN$ 3/3 \\
\bottomrule
\end{tabular}
\end{table}

Under the shifts of the ladder \cref{tab-fashion} repeats \cref{tab-graded-v7}. At the zoom $2$, $\GCN$ keeps its accuracy at the training scale, $83.7\%$ on average, to within $0.1$ points in each split, and $\SNT$ keeps $83.7\%$ of $83.9\%$. The same network with equal grades falls from $81.4\%$ to $52.6\%$, and to $41.9\%$ at the zoom $0.5$; at the zoom $2$, $\GCN$ is right and $\UCN$ wrong on $1{,}597$, $1{,}828$ and $1{,}836$ images in the three splits, and the reverse holds on $190$, $200$ and $204$. The size-normalized network is flat across the zooms, at $70.0$--$74.8\%$, nine points below $\GCN$ at the training scale.

The other zooms show two effects that digits hide. The first is the border of the canvas. Placed on the canvas $57$ at the zoom $1$, the sharp test images are classified $5.9$ points worse by $\GCN$, $6.2$ by $\SNT$ and $6.8$ by $\UCN$ than on the canvas $29$ (the first column of the right block), and on split $1$ the whole loss is already present on a canvas of $33$ pixels, while the accuracy of the size-normalized network does not change. We attribute this to the border: a garment on the canvas $29$ has jets at positions outside the canvas, which the network did not see in training and which every larger canvas supplies, and the zoom $2$ on the canvas $57$ reproduces the training configuration exactly (\cref{sub-impl}). The loss of eight points at the zoom $0.5$, where the reduced image has a margin on the canvas $29$, is of this size. The second effect is the cost of the zooms between the rungs, at which the invariance is not exact. At the zooms $0.7$ and $1.4$, $\GCN$ loses $17.8$ and $18.1$ points and $\SNT$ $20.5$ and $19.8$, against less than three on digits; about six of these points are to be expected from the border alone. At these zooms the size-normalized network is about eight points more accurate than $\GCN$, and $\GCN$ is more accurate than $\SNT$ in two of the three splits. Along the zoom orbits the predicted label is constant for $40.5$--$47.5\%$ of the images for $\GCN$, for $45$--$51.5\%$ for $\SNT$ and for $6.5$--$8.5\%$ for $\UCN$.

The right block of \cref{tab-fashion} tests \cref{prop-blur} on the canvas $57$, as in \cref{tab-blur-v7}. Averaged over the three splits, $\GCN$ loses $3.1$ points at $\sigma_b=1$, $7.0$ at $\sigma_b=1.5$ and $14.8$ at $\sigma_b=2$, and the scale-normalized twin loses $3.4$, $10.0$ and $23.6$. At $\sigma_b=2$, $\GCN$ keeps $63.0\%$ against $54.1\%$; it is right and $\SNT$ wrong on $711$, $399$ and $962$ images in the three splits, the reverse holds on $292$, $281$ and $159$, and $\GCN$ is the more accurate in each split ($p<10^{-4}$ after the Holm adjustment). On half-size images blurred by one pixel it keeps $61.0\%$ against $52.1\%$ and is the more accurate in two splits, by $796$ against $233$ images and by $951$ against $236$, with no significant difference in the remaining one ($317$ against $255$, $p=0.07$). On sharp images at the zooms $1$ and $2$ the two networks do not differ significantly in any split, so that the advantage of $\SNT$ on sharp digits does not recur. The equal-grades twin, which is not invariant, is here almost as robust to blur at the training scale as $\GCN$, with $59.3\%$ at $\sigma_b=2$ and a significant difference in two of the three splits; on digits it was $12$ points behind.

Both comparisons thus recur on a second dataset, with the networks and the protocol unchanged: equal grades lose the invariance under the shifts of the ladder, and the retraction is more robust to a strong blur than scale normalization, by nine points where it was $22$ on digits. The accuracies are lower throughout, and the zooms between the rungs and the border of the canvas cost far more than on digits.

\subsection{Lindeberg's network}\label{sub-exp-gdn}
The pooling of \cref{sub-readout} does not use the grading, and it applies to any network that computes features on a ladder of scales. We apply it to our reimplementation of Lindeberg's Gaussian derivative network \cite{Lindeberg2022}: six layers of learned combinations of scale-normalized Gaussian derivatives, run in eight scale channels, with $20{,}674$ parameters. The network is trained on $6{,}000$ digits for six epochs with its binary cross-entropy loss, once with Lindeberg's maximum over the scale channels and once with our pooling, which weights the channels by their scale relative to the intensity radius of the image; nothing else changes. \Cref{tab-gdn} uses two test sets: $800$ test digits at sizes from $\frac12$ to $4$, and the first $2{,}000$ of the blurred test digits of \cref{tab-blur-v7}. Replacing the maximum over the scale channels by the image-frame pooling reduced the variance of the output over the sizes in $[1/\sqrt2,2\sqrt2]$ from $0.122$ to $0.003$. On the first set, at half, twice and four times the training size, the accuracy rises from $69.3$, $84.1$ and $86.8\%$ to $95.9$, $98.0$ and $97.5\%$. Under blur at the training size both versions are robust, with $89.15$ and $91.45\%$ at $\sigma_b=2$ on the second set; their later layers smooth further (\cref{sub-blur}). For the half-size digits of the second set the pooling is again decisive: $66.25$ against $96.15\%$ for sharp digits, and $73.35$ against $83.65\%$ for digits blurred by one pixel.

The networks with our pooling are then trained again with everything fixed except the first layer, which is graded. Its input, the scale-normalized $2$-jet $(f_1,f_2)$ of the image in each scale channel, is replaced by the retraction $(f_1/\rho,\,f_2/\rho^2)$ with $\rho=(\norm{f_1}^2+\norm{f_2})^{1/2}$, which is $\pi_{\bq}$ for $N=2$, and the four products of frequency zero $\abs{z_1}^2$, $\abs{h_2}^2$, $h_0$ and $\operatorname{Re}(\bar h_2z_1^2)$ of the retracted jet (\cref{sub-forms}) are appended to it, which adds $48$ parameters. The later layers are unchanged and remain scale-normalized. Since $\pi_{\bq}\circ\nu=\pi_{\bq}$ (\cref{thm-forced}(iii)), the graded first layer does not depend on the nominal scale, and by \cref{prop-blur}(ii) and (iii) a blur changes its input only by the change of scale $s\mapsto s+b$ and by the contrast factor of the second normalization.

In the paired counts below the first number counts the digits on which the first network is right and the second wrong, and the second number the reverse; the $p$-values are Holm-adjusted over the six conditions of \cref{tab-gdn}. On sharp digits, under blurs up to $\sigma_b=1.5$ and at the zoom $\frac12$ the graded and the ungraded network are equally accurate, with no significant difference. Under the stronger blurs the graded network is more accurate: $93.10$ against $91.45\%$ at $\sigma_b=2$ ($85$ against $52$ digits, $p=0.03$), and $89.75$ against $83.65\%$ for half-size digits blurred by one pixel ($181$ against $59$, $p<10^{-14}$). The comparison replicates in a second version of the network with the four scale channels $\sigma_0\in\{2^{-1/2},2^{1/2},2^{3/2},2^{5/2}\}$ in place of eight, trained in the same way: $92.50$ against $90.50\%$ at $\sigma_b=2$ ($93$ against $53$, $p=0.006$) and $88.35$ against $82.25\%$ for half-size digits blurred by one pixel ($215$ against $93$, $p<10^{-10}$), with no other significant difference. On the size test the grading changes the accuracy in neither version beyond the sampling error of $800$ digits, and the variance of the output across sizes, $0.004$ and $0.009$, stays far below the $0.122$ of the maximum over channels. Each network was trained once, on the same training digits, so the two versions replicate the comparison across architectures, not across training data, and the paired tests account for the variation over test digits, not for that between training runs.

The gain comes from the retraction. A third version with eight scale channels retracts the jet of the first layer without appending the products, and so has exactly the parameters of the ungraded network. It is the most robust of all: $97.30$, $95.10$ and $91.45\%$ at $\sigma_b=1.5$, at $\sigma_b=2$ and for half-size digits blurred by one pixel, against $96.10$, $91.45$ and $83.65\%$ for the ungraded network ($45$ against $21$, $104$ against $31$ and $200$ against $44$ digits; $p=0.017$, $p<10^{-8}$ and $p<10^{-23}$), with no significant difference on sharp digits, at $\sigma_b=1$ or at half size. The products lower the gain slightly: against the retraction alone, the first layer with products is right and the other wrong on $37$ digits against $77$ at $\sigma_b=2$ ($p=0.001$), and on $65$ against $99$ for the blurred half-size digits ($p=0.049$).

\begin{table}[htbp]
\centering\scriptsize
\caption{Lindeberg's Gaussian derivative network \cite{Lindeberg2022} in our reimplementation, in versions with eight and with four scale channels, each trained once with the same architecture, the same $6{,}000$ digits at size $1$, six epochs and the same loss. The pooling is Lindeberg's maximum over the scale channels or the image-frame pooling of \cref{sub-readout}; the graded first layer retracts the scale-normalized $2$-jet and appends its four frequency-zero products ($20{,}722$ parameters against $20{,}674$), and the retraction-only version retracts it without the products ($20{,}674$ parameters) (\cref{sub-exp-gdn}). Upper panel: accuracy (\%) on $800$ test digits at the sizes shown (canvas $113$ above $\sqrt2$) and mean variance of the normalized outputs over the nine sizes in $[1/\sqrt2,2\sqrt2]$. Middle panel: accuracy (\%) on a second test set, the first $2{,}000$ test digits of \cref{tab-blur-v7}, blurred by $\sigma_b$ input pixels, and at the zoom $\frac12$ sharp and with $\sigma_b=1$; on $2{,}000$ digits every accuracy is a multiple of $0.05$ and is given exactly. Lower panel: the digits of the second set on which only the first or only the second network of each pair is right, and the exact two-sided McNemar test between them, Holm-adjusted over the six conditions \cite{Holm1979}.}\label{tab-gdn}
\setlength{\tabcolsep}{2.6pt}
\begin{tabular}{@{}lrrrrrrrr@{}}
\toprule
Pooling & $0.5$ & $\frac1{\sqrt2}$ & $1$ & $\sqrt2$ & $2$ & $2\sqrt2$ & $4$ & var. \\
\midrule
\multicolumn{9}{@{}l}{\emph{eight scale channels}} \\
\quad maximum over channels \cite{Lindeberg2022} & 69.3 & 96.6 & 98.0 & 91.4 & 84.1 & 81.6 & 86.8 & 0.122 \\
\quad image-frame pooling & 95.9 & 98.1 & 98.5 & 98.5 & 98.0 & 97.9 & 97.5 & 0.003 \\
\quad image-frame pooling, graded first layer & 96.8 & 98.3 & 98.8 & 98.8 & 98.8 & 98.1 & 98.1 & 0.004 \\
\quad image-frame pooling, retraction only & 96.4 & 98.9 & 98.6 & 98.5 & 97.9 & 97.8 & 97.9 & 0.004 \\
\multicolumn{9}{@{}l}{\emph{four scale channels}} \\
\quad image-frame pooling & 96.4 & 98.1 & 98.6 & 98.3 & 99.1 & 98.4 & 97.6 & 0.005 \\
\quad image-frame pooling, graded first layer & 96.6 & 98.0 & 98.4 & 98.5 & 97.9 & 97.8 & 97.4 & 0.009 \\
\bottomrule
\end{tabular}

\medskip
\begin{tabular}{@{}lrrrrrr@{}}
\toprule
& \multicolumn{4}{c}{blur $\sigma_b$, zoom $1$} & \multicolumn{2}{c}{zoom, $\sigma_b$} \\
\cmidrule(lr){2-5}\cmidrule(l){6-7}
Pooling & $0$ & $1$ & $1.5$ & $2$ & $\frac12$, $0$ & $\frac12$, $1$ \\
\midrule
\multicolumn{7}{@{}l}{\emph{eight scale channels}} \\
\quad maximum over channels \cite{Lindeberg2022} & 97.60 & 96.70 & 94.80 & 89.15 & 66.25 & 73.35 \\
\quad image-frame pooling & 98.45 & 98.05 & 96.10 & 91.45 & 96.15 & 83.65 \\
\quad image-frame pooling, graded first layer & 98.25 & 97.85 & 96.60 & 93.10 & 96.25 & 89.75 \\
\quad image-frame pooling, retraction only & 98.15 & 98.20 & 97.30 & 95.10 & 96.45 & 91.45 \\
\multicolumn{7}{@{}l}{\emph{four scale channels}} \\
\quad image-frame pooling & 98.20 & 97.95 & 96.50 & 90.50 & 96.90 & 82.25 \\
\quad image-frame pooling, graded first layer & 98.20 & 97.50 & 96.15 & 92.50 & 96.50 & 88.35 \\
\midrule
\multicolumn{7}{@{}l}{\emph{eight scale channels}} \\
\quad graded vs.\ ungraded, digits right only for each & 10/14 & 15/19 & 35/25 & 85/52 & 32/30 & 181/59 \\
\quad graded vs.\ ungraded, $p$ (Holm) & 1.00 & 1.00 & 0.98 & 0.030 & 1.00 & $8\cdot10^{-15}$ \\
\quad retraction only vs.\ ungraded, digits right only for each & 8/14 & 16/13 & 45/21 & 104/31 & 31/25 & 200/44 \\
\quad retraction only vs.\ ungraded, $p$ (Holm) & 0.86 & 1.00 & 0.017 & $1\cdot10^{-9}$ & 1.00 & $4\cdot10^{-24}$ \\
\quad graded vs.\ retraction only, digits right only for each & 14/12 & 15/22 & 21/35 & 37/77 & 29/33 & 65/99 \\
\quad graded vs.\ retraction only, $p$ (Holm) & 1.00 & 0.97 & 0.33 & 0.001 & 1.00 & 0.049 \\
\multicolumn{7}{@{}l}{\emph{four scale channels}} \\
\quad graded vs.\ ungraded, digits right only for each & 12/12 & 8/17 & 29/36 & 93/53 & 25/33 & 215/93 \\
\quad graded vs.\ ungraded, $p$ (Holm) & 1.00 & 0.43 & 1.00 & 0.006 & 1.00 & $2\cdot10^{-11}$ \\
\bottomrule
\end{tabular}
\end{table}

Lindeberg's network differs from $\GCN$ in depth, training data, number of epochs and loss, so these numbers concern the pooling and the grading within that architecture and do not compare the two architectures. Under its own protocol the ungraded version is as robust to blur at the training size as $\GCN$, with $89.15\%$ against $88.90\%$ at $\sigma_b=2$ on the same $2{,}000$ digits, and more accurate on sharp digits. In both architectures the retraction supplies the robustness to blur that scale normalization lacks: by a wide margin in the shallow network of \cref{sub-exp-blur}, where no later smoothing compensates, and by a smaller but significant one in Lindeberg's deeper network, whose later layers compensate in part.

\section{Conclusion}\label{sec-conclusion}
Derivative order grades the Gaussian jet of an image, and invariance under dilations forces this grading: every invariant local feature factors through the scale-normalized jet or, if it does not use the scale, through the retraction onto the weighted sphere of the grading. The two realizations differ under an unknown Gaussian blur: every function of the raw jet, the retraction included, sees it as a change of scale, while the normalization by the nominal scale adds a factor that depends on the derivative order. Near the singular point of the retraction the ideal graded quotient separates, with a uniform margin, two image families on which the score gap of every radial probe classifier and of every classifier continuous at their common limit tends to zero. For the ideal network with $\kappa=2$, \cref{prop-equiv} gives exact invariance under the dilations by powers of two when the sampling grids match and the pooling bins lie inside the ladder of scales. The implementation uses a fixed positive regularization of the norm, $\varepsilon=10^{-12}$, which breaks the exact homogeneity of the retraction and loses the uniform separation of \cref{thm-local-separation} on the constructed family (\cref{prop-regularized-collapse}). On MNIST the implemented network attained mean accuracies of $95.4\%$, $96.0\%$ and $96.0\%$ at the zooms $0.5$, $1$ and $2$, and $88.4\%$ under a Gaussian blur of two pixels, against $66.2\%$ for the same network on scale-normalized derivatives; on Fashion-MNIST both comparisons recur with smaller margins; and the retraction alone, applied in the first layer of Lindeberg's Gaussian derivative network, makes that network more robust to blur as well.

The scale-normalized network is more accurate than $\GCN$ on sharp images, by $0.8$ to $1.3$ points on average over the three splits and by up to $2.6$ in one, by up to $0.9$ points on MNIST-Scale, and under mild blur ($\sigma_b\le0.5$, or the resolution halved). A possible cause, which we have not tested, is the conditioning of the retraction near vanishing blocks, where it is only H\"older continuous (\cref{lem-regularity}). The invariance of the pooled features is exact for the even shifts of the ladder and not for continuous zoom: at the zooms $0.7$ and $1.4$, $\GCN$ loses less than three points on digits and about eighteen on Fashion-MNIST (\cref{sub-exp-fashion}). By \cref{prop-blur} the grading factor that affects the scale-normalized network is largest at the finest rungs, where $\mu^2=s/(s+b)$ is farthest from one; a scale-normalized network whose ladder starts at a coarser scale would be less sensitive to a blur of a given size, at the price of the fine scales that resolve the strokes, and we have not trained such a network. Deeper networks on scale-normalized derivatives, such as Lindeberg's, recover much of the robustness through their later layers, and retracting their first layer adds to it. Training with blur augmentation could supply part of the robustness, as scale augmentation supplies part of the scale invariance; $\GCN$ needs neither.

The experiments use two datasets of isolated, centred objects on images of $28$ pixels, three training splits for the shallow networks and one training run for each version of Lindeberg's network, and a Gaussian model of blur. Real defocus is not Gaussian, and invariants to Gaussian blur in the sense of \cite{FlusserEtAl2016, FlusserEtAl2023} are a different and complementary tool. The pooling uses a global intensity frame, which isolated digits provide; images with several objects or with clutter are not tested, and neither are the scale-equivariant networks with learned filters \cite{WorrallWelling2019, Sosnovik2020}. It remains to test the graded retraction on natural images, on images with clutter and under real optical blur.

\appendix

\section{Discretization and numerical verification}\label{app-arch}

\subsection{Discretization}\label{sub-impl}
Images are placed on odd square canvases, of $29$ pixels for MNIST and $57$ for images enlarged beyond the canvas, and upsampled by two with bicubic interpolation aligned at the corners, so that pixel $x$ of the input is pixel $2x$ of the working grid. The ladder $s\in\{2,4,\dots,4096\}$ in working pixels$^2$, twelve rungs one octave of $s$ apart, corresponds to $\sigma=\sqrt s/2\in\{0.71,1,1.41,\dots,32\}$ input pixels; the coarse rungs cost almost nothing and give the bins room on both sides. The Gaussian derivatives are separable sampled kernels truncated at $4\sqrt s$, in the cross-correlation form $K_a(t)=(\partial^ag_s)(-t)$. The jet at the rung $s_i$ is sampled on a pyramid with stride $2^{\lfloor i/2\rfloor+1}$ working pixels, that is $2,2,4,4,\dots,64,64$, on the grid $x=c+\mathrm{stride}\cdot t$ through the canvas centre $c$. This grid is symmetric under the rotations by $90^\circ$ and nests exactly under the dilation by two from the canvas $C$ to the canvas $2C-1$. A dilation by $\sqrt2$ shifts the ladder by one rung but resamples the pyramid, so the exact statements of \cref{prop-equiv}(iii) are for even shifts. Positions whose foreground gate is at most $0.02$, or whose bin memberships sum to at most $10^{-3}$, are skipped; since $g$ and $m_b$ are invariant, so is the skipped set, but the selection is discontinuous where a threshold is crossed. The homogeneous norm uses $\norm{f_k}^2+\varepsilon^2$ with $\varepsilon=10^{-12}$, so that the exactly flat background is mapped to zero, and it is computed in the log domain, $\log\rhot=\frac12\logsumexp_k\bigl(\frac{2}{\theta_k}\log\norm{f_k}\bigr)$. For $\SNT$ the harmonic coordinates of order $k$ are multiplied by $s^{k/2}$ before the radial projection. One epoch of $10^4$ images takes about $35$ seconds on two CPU cores.

\subsection{Numerical verification}\label{sub-verification}
The statements of \cref{sec-jets,sec-conv,sec-invariants,sec-blur} and of \cref{thm-extrapolation} were checked with the scripts \texttt{verify\_identities}, \texttt{verify\_blur} and \texttt{verify\_v6} of the code \cite{ShaskaCode2026}. Symbolically: the statements of \cref{thm-forced}, together with the realizability of the jets of order at most $3$ (the Gram matrix of the Gaussian derivatives of order at most $3$ has full rank); \cref{prop-blur} and the example that follows it; the two identities of \cref{lem-octave} on polynomial images, for which the jet is finite; and the identities of \cref{sub-forms} and of the transvectant $(f_2,f_2)^{(2)}$. Numerically, for \cref{thm-extrapolation}: the activation pattern of a random ReLU network is constant beyond a computable $\mu_0$, where the network equals its cubic exactly, and the radially projected network converges at the rate $\mu^{-1}$. On real digits the sampled jets follow \cref{sec-blur} up to discretization: for blurs of $0.5$ and $1$ input pixel, between the blurred digit at $s\in\{4,8,16\}$ working pixels$^2$ and the sharp digit at $s+b$, the retracted jets agree to $0.2$--$1.3\%$ of their size, the sampled Gaussian not being an exact semigroup, while the radial projections of the normalized jets differ by $2$--$24\%$, most at the finest rung.

The equivariance statements were verified numerically (\cref{tab-verification}). The position features are invariant under their own grading action to $10^{-15}$, and under the exact dilation by two of real digits the features of $\GCN$ at rung $i+2$ agree with those at rung $i$ to $0.01$--$0.24$ standard deviations across digits, those of $\SNT$ to at most $0.08$, while those of $\UCN$ differ by up to $1.86$. Between $s=4$ and $s'=6$ working pixels$^2$ the normalized jet of order $0$ and $1$ of a digit changes by $13\%$ and $20\%$ of its size. The grading factor of \cref{lem-octave} alone leaves $13\%$ at order $0$, where it is trivial, and overshoots to $33\%$ at order $1$; the grading factor and the heat correction, truncated at the $5$-jet, leave $1\%$ and $5\%$, which combine the discretization of the derivatives with the truncation of the heat series (E6).

\begin{table}[htbp]
\centering\footnotesize
\caption{Numerical verification (\texttt{test\_equivariance.py}); E3, E4 in double precision. Relative errors are maxima over random inputs or over $16$--$32$ MNIST digits; ``std'' denotes standard deviations of the feature across digits. E1, E2, E5 and E6 measure the discretization of the image-level statements; only E3 and E4 are at machine precision.}\label{tab-verification}
\begin{tabular}{@{}p{0.555\textwidth}p{0.415\textwidth}@{}}
\toprule
Check & Result \\
\midrule
E1\ sampled Gaussian derivatives vs.\ the analytic jet of a blob, $s=1,2,4,8$ & $3.9\cdot10^{-3}, 1.5\cdot10^{-3}, 3.6\cdot10^{-3}, 1.2\cdot10^{-3}$ \\
E2\ scale-space covariance \eqref{eq-dil} under $\Dil_2$, analytic images, $s=1,2,4$ & $7.0\cdot10^{-5}, 1.2\cdot10^{-5}, 3.1\cdot10^{-5}$ \\
E3\ position features under $\mu\star_{\btheta}$ for $\btheta=(1,2,3)$ / $(1,1,1)$ / $(3,1,2)$ & $1.5\cdot10^{-15} / 1.1\cdot10^{-15} / 1.3\cdot10^{-15}$ \\
E3\ features of $\btheta=(1,1,1)$ / $(3,1,2)$ under the dilation action $\mu\star_{\bq}$ & 0.95 / 1.14 \\
E4\ harmonic coordinates transform by $e^{ij\phi}$; invariant readout rotation-invariant & $2.6\cdot10^{-16}$; $6.9\cdot10^{-16}$ \\
E4\ pooled vector, invariant readout, under exact $90^\circ$ rotations of digits, $\GCN$ / $\UCN$ & $3.1\cdot10^{-14}$ / $1.3\cdot10^{-14}$ \\
E5\ per-rung features under $\Dil_2$ of digits, rung $i+2$ vs.\ $i$, ten rungs (std), $\GCN$ & 0.24, 0.06, 0.05, 0.03, 0.04, 0.02, 0.02, 0.01, 0.03, 0.02 \\
E5\ same, $\UCN$ & 1.86, 1.14, 0.70, 0.38, 0.28, 0.12, 0.08, 0.04, 0.85, 0.84 \\
E5\ same, wrong grading $(3,1,2)$ & 2.58, 0.95, 0.55, 0.21, 0.13, 0.06, 0.06, 0.03, 0.11, 0.06 \\
E5\ same, scale-normalized twin $\textup{SNT}$ & 0.08, 0.07, 0.06, 0.03, 0.05, 0.02, 0.03, 0.01, 0.01, 0.00 \\
E6\ \eqref{eq-octave} on digits, $s=4\to6$, order $0$ / order $1$: no correction, $\mu\star$ only, $\mu\star$ and heat & $1.3\cdot10^{-1}$, $1.3\cdot10^{-1}$, $9.8\cdot10^{-3}$ / $2.0\cdot10^{-1}$, $3.3\cdot10^{-1}$, $5.4\cdot10^{-2}$ \\
\bottomrule
\end{tabular}
\end{table}

\section*{Declarations}
\subsection*{Funding} No funding was received for this work.

\subsection*{Competing interests} The author has no competing interests to declare.

\subsection*{Data availability} The MNIST and Fashion-MNIST images are available from their original sources and are not redistributed. The result files from which every table and figure of this paper is generated are openly available on Zenodo at \url{https://doi.org/10.5281/zenodo.23997172} \cite{ShaskaCode2026}.

\subsection*{Code availability} The code and the trained networks are deposited with the result files \cite{ShaskaCode2026}, with instructions for regenerating each table from the stored results and for retraining every network.

\end{document}